\documentclass[twocolumn,final]{svjour3}          
\usepackage{inputenc}
\usepackage{graphicx}
\usepackage{cite}
\usepackage{amssymb,amsfonts}
\usepackage{caption}
\usepackage{textcomp}
\usepackage{url}
\usepackage{multirow}
\usepackage{comment} 
\usepackage{diffs,xspace}
\usepackage{algorithm}
\usepackage{adjustbox}
\usepackage{placeins}
\usepackage{algpseudocode}
\usepackage{subfig}
\usepackage{xcolor}
\usepackage[round]{natbib} 

\RequirePackage[centertags]{amsmath}
\usepackage{cuted}
\usepackage{array,booktabs}
\usepackage{makecell,tabularx} 
\setcellgapes{2pt}
\usepackage[nointegrals]{wasysym}

\usepackage{balance}
\usepackage{hyperref}
\hypersetup{
	colorlinks=true,
	linkcolor=blue,
	citecolor=blue,
	filecolor=magenta,      
	urlcolor=blue,
}
\begin{document}

\title{Failure Prediction in Production Line Based on Federated Learning: An Empirical Study
}


\author{Ning Ge \and
        Guanghao Li \and
        Li Zhang \and
        Yi Liu}


\institute{Ning Ge \at
           School of Software, Beihang University, Beijing, China \\
           \email{gening@buaa.edu.cn}  \\
           \emph{She was also with the Key Laboratory of Safety-Critical Software (Nanjing University of Aeronautics and Astronautics), Ministry of Industry and Information Technology.}
           \and
           Guanghao Li \at
           School of Software, Beihang University, Beijing, China \\
           \email{liguanghao@buaa.edu.cn}
           \and
           Li Zhang \at
           School of Computer Science and Engineering, Beihang University, Beijing, China \\
           \email{lily@buaa.edu.cn} \\
           \emph{Li Zhang is the corresponding author.}
           \and
           Yi Liu\at
           School of Computer Science and Engineering, Beihang University, Beijing, China \\
           \email{zy1906505@buaa.edu.cn}
}

\date{Received: date / Accepted: date}

\maketitle

\begin{abstract}
Data protection across organizations is limiting the application of centralized learning (CL) techniques. Federated learning (FL) enables multiple participants to build a learning model without sharing data. Nevertheless, there is very few research works on FL in intelligent manufacturing. This paper presents the results of an empirical study on failure prediction in the production line based on FL. This paper (1) designs Federated Support Vector Machine (FedSVM) and Federated Random Forest (FedRF) algorithms for the horizontal FL and vertical FL scenarios, respectively; (2) proposes an experiment process for evaluating the effectiveness between the FL and CL algorithms; (3) finds that the performance of FL and CL are not significantly different on the global testing data, on the random partial testing data, and on the estimated unknown Bosch data, respectively. The fact that the testing data is heterogeneous enhances our findings. Our study reveals that FL can replace CL for failure prediction. 

\keywords{Empirical study \and Federated learning \and Failure prediction \and Production line \and Manufacturing \and Bosch dataset}

\end{abstract}

\section{Introduction}
\label{sec:intro}

Artificial intelligence (AI) is one of the core techniques in the fourth industrial revolution \cite{zhong2017intelligent,li2017applications}. For instance, AI techniques have been employed to improve the early detection or fault prediction within production lines \cite{tao2018data,kusiak2017smart} in intelligent manufacturing (IM). A prominent achievement is that in 2012, Intel saved \$3 million in manufacturing costs through the use of predictive analytics to prioritize its silicon chip inspections \cite{ronenutilizing}. To this end, the Bosch company has published a dataset of product quality prediction on the data analysis competition platform Kaggle\footnote{https://www.kaggle.com/c/bosch-production-line-performance}. The dataset reflects the relevant parameters and equipment operation of each product during the production process, hoping to reduce defective products. By using this dataset, several studies have proposed product quality prediction methods based on centralized learning (CL) algorithms ~\cite{carbery2019new,carbery2018bayesian,zhang2016predict,khoza2019comparing,kotenko2019improving,mangal2016using,hebert2016predicting,maurya2016bayesian,huang2019enhancing,moldovan2019time,liu2020adversarial}. The results of these work revealed two important facts. First, data preprocessing on the Bosch dataset has a significant impact on the prediction result. Second, among the CL algorithms, the Random Forest algorithm performed better than others when time-series features are excluded \cite{zhang2016predict,khoza2019comparing} and the Long Short Term Memory (LSTM) network can improve the prediction result when time-series features are included \cite{huang2019enhancing}. CL methods usually require that clients upload data to the server. The server trains an integrated model based on the shared data. However, in real-world scenarios, private data belongs to multiple independent organizations and cannot be shared with others. Therefore, data sharing has arisen to be an obstacle to the application of CL methods. As UC Berkeley pointed out in a report published in 2017, learning from private data is one of the challenges encountered by the current AI field \cite{stoica2017berkeley}. 


In response to the above problem, Google proposed the concept of federated learning (FL) in 2016 to decentralize the collaborative learning  \cite{mcmahan2016communication}. FL trains models across multiple decentralized devices holding local private data samples, without exchanging or sharing these samples. Clients then encrypt and transmit the trained model or parameters back to the server to build an integrated model. On the one hand, FL addresses critical issues such as data privacy and data security \cite{yang2019federated,li2020federated}. On the other hand, as FL trains model locally without removing the private data features, the data samples can be more complete than CL. FL has received widespread attention recently and has been applied in many fields. For example, Google launched a project on the prediction of mobile keyboard input in 2018 and obtained better prediction results using FL \cite{hard2018federated}. Intel started to support the hardware architecture for  FL\footnote{https://www.intel.com/content/www/us/en/artificial-intelligence/posts/federated-learning-for-medical-imaging.html}. FL has also been applied to various fields like medical systems \cite{sheller2018multi,brisimi2018federated,boughorbel2019federated,huang2019patient}, Internet finance \cite{suzumura2019towards,yang2019ffd}, smart city \cite{samarakoon2019distributed,saputra2019energy,liu2020privacy}, edge computing \cite{bakopoulou2019federated}, IOT \cite{nguyen2019diot,chen2020federated}, Cyber Physical Systems (CPS) \cite{mowla2019federated,aussel2020combining}, etc. 

In the field of IM, the manufacturing process often crosses enterprises, workshops, or even production lines. Due to data privacy and security, the application of CL is becoming the bottleneck. The studies on the failure prediction in the production line have been conducted under the assumption that all data are shared. Under the premise of data protection between production lines, these methods become unfeasible. Nevertheless, according to our investigation \cite{liu2020systematic}, applying FL to the field of IM is still in its infancy, and there are very few research works on it. Although FL has achieved success in many fields, whether it can achieve similar performance and replace CL in real-world manufacturing is still an open question. Accordingly, this paper bridges this gap by reporting an empirical study of defective product prediction in the production line by comparing FL with CL. This study is the first attempt to design FL algorithms for failure prediction in the production line. 

In addressing our research goal, we construct a horizontal FL (HFL) scenario and a vertical FL (VFL) scenario, respectively. In the former, we compared the Federated Support Vector Machine (FedSVM) and the Support Vector Machine (SVM) algorithms. In the latter, we compare the Federated Random Forest (FedRF) and the Random Forest (RF) algorithms. Each group of contrast experiment focuses on four research questions (RQs).
\begin{itemize}
	\item[$\circ$]
	\emph{RQ1}, we are interested in whether FL can replace CL on the whole Bosch testing data. Hence, we ask whether the average performance of FedSVM (FedRF resp.) is similar to that of SVM (RF resp.) on the whole Bosch testing data? 
	\item[$\circ$]
	\emph{RQ2}, we are interested in whether FL can replace CL on part of the given Bosch testing data. Therefore, we ask whether the average performance of FedSVM (FedRF resp.) is similar to that of SVM (RF resp.) on random partial Bosch testing data?
	\item[$\circ$] 
	\emph{RQ3}, if FL is applied to predicting defective products on other unknown Bosch data, whether FL can replace CL. Hence, we ask whether the average performance of FedSVM (FedRF resp.) is similar to that of SVM (RF resp.) on the estimated unknown Bosch testing data? 
	\item[$\circ$] 
	\emph{RQ4}, we want to know whether the Bosch testing data is heterogeneous or not. If the data is heterogeneous and the answers of RQ1-RQ3 are positive, it means that the selected FL algorithms are robust. Therefore, we ask, is there local heterogeneity within the given testing data?
\end{itemize}

Answers to these questions are helpful for manufacturing as they provide insights into how to evaluate an FL algorithm can achieve similar effects as a CL one in a real-world application, and then replace it. To answer these questions, this paper makes methodological, substantive, and theoretical contributions to the literature on failure prediction in the production line. 

First, in terms of algorithm design, to conduct empirical research on the failure prediction based on HFL and VFL, we first select SVM and RF from previous works as CL baseline because both algorithms performed better than others in the previous works. Then, we design FedSVM and FedRF algorithms for our problem based on existing works \cite{bakopoulou2019federated,liu2020federated}. In terms of novelty in the algorithm design, our work improved FedRF by introducing optimal feature selection and pruning steps. 

Second, we propose a set of experimental methodology to compare the average performance of FL and CL in manufacturing from three aspects: on the whole testing data, on the random partial testing data, and on the estimated unknown testing data. To compare the performance of FL and CL on the estimated unknown Bosch data, we follow the process of Measurement Systems Analysis (MSA) to fit a Markov process model of prediction error $\mathtt{M_F}$ ($\mathtt{M_C}$ resp.) based on FL (CL resp.) and then compare the two fitting models. The experimental methods also assess the heterogeneity of the testing data to enhance the results of RQ1-RQ3 on whether FL can replace CL. 

Last, our empirical research shows that our FL solution is not significantly different from the CL solution for failure prediction, on the whole, on the random partial, and on estimated unknown Bosch testing data. FL can replace CL in the application of failure prediction within the manufacturing process. 

In the remainder of this paper, we first introduce the background knowledge of our empirical research and related work in Section \ref{sec:background}, present the designed HFL and VFL algorithms in Section \ref{sec:fl}, and outline the experimental methodology in Section \ref{sec:methodo}. Section \ref{sec:exp} presents the results of our empirical study. Section \ref{sec:threats} discusses the threat of validity. Section \ref{sec:discussion} discusses whether FL can replace CL. Section \ref{sec:conclu} concludes this paper and looks forward to future work.

\section{Background and Related Work}
\label{sec:background}

This section introduces previous works on Bosch product quality prediction based on the CL algorithms and discusses the necessity of conducting empirical research on the FL algorithms for this problem; and then, briefly explains the principles of federated learning.

\subsection{Failure Prediction in Bosch Production Line}

Bosch is one of the worldwide leading manufacturing companies. It ensures  high quality of the production by monitoring its parts in the  manufacturing processes. Because Bosch records detailed data for each step on the assembly lines, they can apply advanced techniques to improve the manufacturing processes. To this end, Bosch has published a dataset on the Kaggle competition platform to predict internal failures by thousands of measurements and tests made for each component along the assembly line. Some studies have analyzed the dataset and carried out approaches to predicting product quality based on CL algorithms excluding time-series features \cite{carbery2019new,carbery2018bayesian,zhang2016predict,khoza2019comparing,kotenko2019improving,mangal2016using,hebert2016predicting,maurya2016bayesian} or including time-series features \cite{huang2019enhancing,moldovan2019time,liu2020adversarial}, as summarized in Table \ref{tab:related}. These works were conducted based on a set of learning methods, including Logistic Regression (LR), Gradient Boosting Machine (GBM), Random Forest (RF), Gradient Boosted Trees (GBT), Naive Bayes (NB), Bayesian Network (BN), K-Nearest Neighbors (KNN), Support Vector Machines (SVM), Multilayer Perceptron Classifier (MPC), Majority Voting (MV), Decision Tree (DT), Statistical Process Control (SPC), etc. 

\begin{table*}[!htbp]
	\centering
	\setlength{\extrarowheight}{2pt}
	\caption{Research Works on Failure Prediction in Bosch Production Line}    		
	\scriptsize
	\begin{tabular}{  p{1.7cm}  c  p{4.6cm}  p{6cm} }
		\toprule
		\textbf{Objectives} & \textbf{Ref} & \textbf{Learning Method}  & \textbf{Contribution} \\
		\hline		
		\multirow{9}{1.7cm}{\textbf{Improving predictive model without time-series features} } 
		& \cite{carbery2018bayesian} &  XGBoost, BN & BN model performs well in failure prediction.\\	
		& \cite{zhang2016predict} & RF, Gradient Boosting, LR, NB, DT & RF performs better on different clusters. \\		
		& \cite{khoza2019comparing} & RF, SVM, NB, SPC & RF outperforms other models. \\			
		& \cite{kotenko2019improving} & SVM, KNN, Perceptron, LR, DT, MV & SVM and MV outperform other models.  \\		
		& \cite{mangal2016using} & LR, Extra Trees Classifier, RF, XGBoost & XGBoost and RF outperform other methods. \\				
		& \cite{hebert2016predicting} & LR, RF, XGBoost & RF and XGBoost can properly identify conditions leading to failure events. \\	
		& \cite{maurya2016bayesian} & XGBoost & Optimize MCC by using GBM as a base classifier. \\			
		\hline	
		\multirow{4}{1.7cm}{\textbf{Improving predictive model with time-series features}}	
		& \cite{huang2019enhancing}  & Ontology-based LSTM neural network & Ontology-based LSTM neural network yields a better performance \\
		& \cite{moldovan2019time} & RF, GBT, NB, KNN, SVM, and MPC & The LSTM RNN  model outperform others. \\	
		& \cite{liu2020adversarial}  & SP-LSTM models &  The A-Bi-SP-LSTM model outperforms other models. \\			
		\bottomrule
	\end{tabular}
	\label{tab:related}
\end{table*}

Carbery et al. \cite{carbery2019new} conducted a systematic feature analysis on the Bosch dataset and used BN to predict product quality \cite{carbery2018bayesian}. Zhang et al. \cite{zhang2016predict} weakened data heterogeneity through clustering, and then applied RF, Boosting, LR, NB, DT to each cluster. They showed that RF performed better than other algorithms. Based on \cite{zhang2016predict}, Khoza et al. \cite{khoza2019comparing} compared RF, NB, SVM, and SPC. They also showed that RF performed better. Kotenko et al. \cite{kotenko2019improving} used SVM, KNN, LR, Perceptron, DT, and MV and showed that SVM achieved relatively higher prediction accuracy. Mangal et al. \cite{mangal2016using} conducted a visual analysis of the three types of data: categorical, numeric, and time-series features on the Bosch dataset, and used LR, Extra Trees Classifier, RF, XGBoost for quality prediction. Hebert \cite{hebert2016predicting} found that RF and XGBoost could properly identify conditions that lead to failure events. Maurya \cite{maurya2016bayesian} optimized MCC measure by using GBM as a base classifier. 

The studies \cite{moldovan2019time,huang2019enhancing,liu2020adversarial} have considered time-series features. Huang et al. \cite{huang2019enhancing} constructed an LSTM network model based on the time-series features of the data, which has great enlightening significance for other researchers. Moldovan et al. \cite{moldovan2019time} found that the LSTM RNN model outperformed other machine learning models. Liu et al. \cite{liu2020adversarial} proposed an end-to-end unified quality prediction framework to capture temporal interactions among the features of different processes. They showed that the A-Bi-SP-LSTM model outperforms the existing data-driven methods. 

Despite these studies on learning features for failure prediction, they have been all conducted under the assumption that all data are shared. Under the premise of data protection between production lines, these methods become unpractical. An open question is whether there exist some FL algorithms that can replace CL methods. Thus, our study is a first attempt to design FL algorithms for failure prediction in the assembly line. 

\subsection{Overview of Federated Learning}

According to the relationship between the datasets provided by the clients, we usually classify FL into HFL, VFL, and federated transfer learning (TFL) \cite{kairouz2019advances}. This article focuses on the HFL and VFL scenarios. In HFL, data are partitioned by user IDs or device IDs. For example, in the context of manufacturing, clients A and B are two independent factories having the same production line structure and machine configuration. The data features in each production line are roughly the same, while the product IDs on their production lines do not overlap. VFL is applicable to the cases that two datasets share the same sample ID space but differ in feature space. For example, production lines A and B are independent, but one belongs to the upstream, and the other belongs to the downstream of the entire production line. Their product IDs are likely to be the same, which guarantees the intersection of their product space. However, since both A and B record part of manufacturing behavior, their feature spaces are very different. In this paper, we design FedSVM as the HFL model, and design FedRF as the VFL model, respectively, to compare with centralized SVM and RF models. 


\section{Federated SVM and Federated RF}
\label{sec:fl}

This section introduces the design of the federation learning algorithms in this work. Through the investigation in Sect. \ref{sec:background}, we selected SVM and RF models, which performed well on the Bosch dataset, as the baseline. In the HFL scenario, we reuse the existing FedSVM algorithm to compare it with SVM. In the VFL scenario, we improve the existing Federated Forest algorithm \cite{liu2020federated} to compare with RF.

\subsection{Federated SVM}

\begin{figure*}[!htbp]
	\centering
	\includegraphics[height=0.35\textheight]{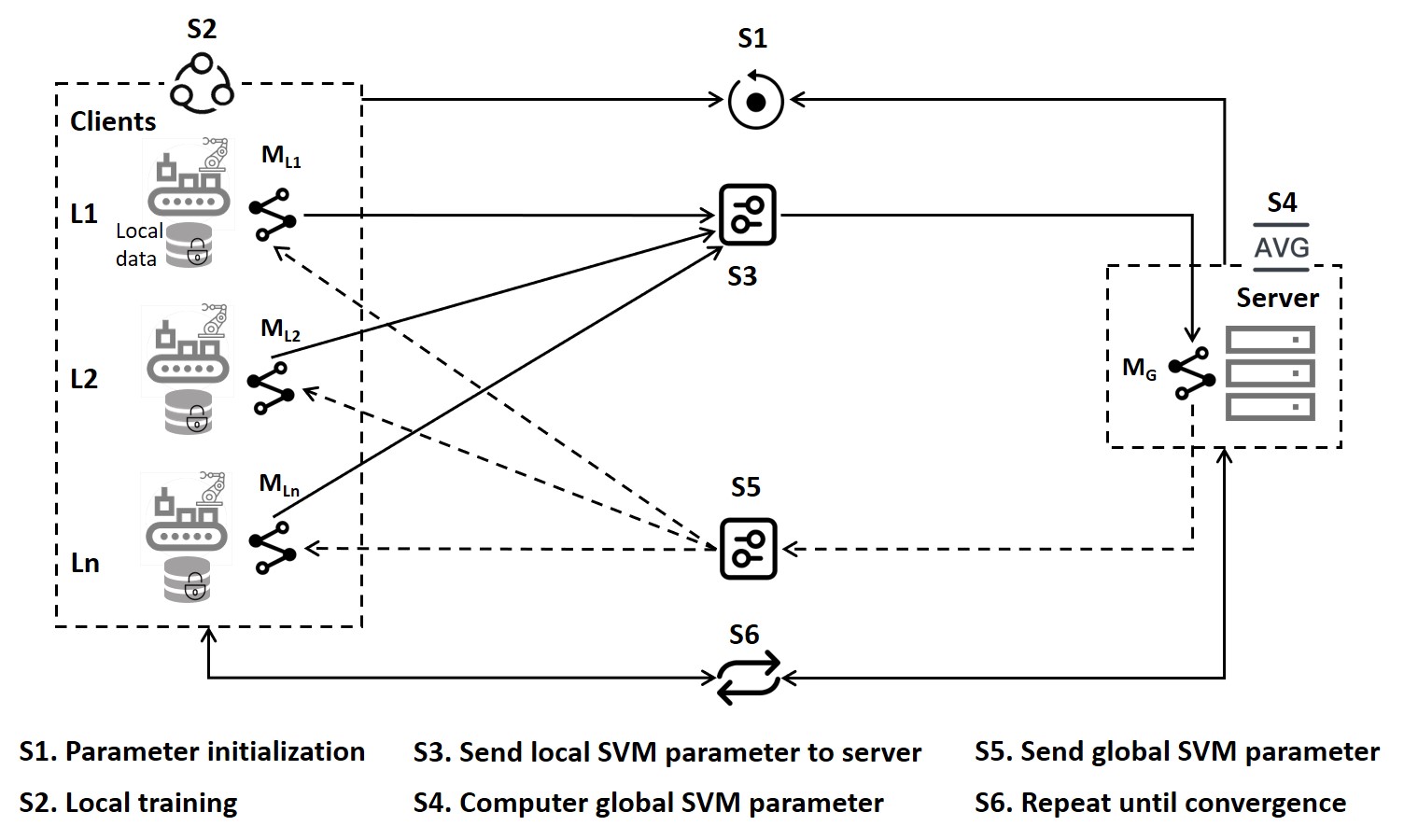}
	\caption{The Training Process of FedSVM}
	\label{fig:FedSVM}
\end{figure*}

As a supervised learning algorithm, SVM is suitable for classification and regression analysis. The effectiveness of SVM mainly depends on how to select 1) the kernel, 2) the kernel's parameters, and 3) the soft margin parameter. In this work, we use FedSVM with a linear kernel. SVM specifies parameters and intercepts, which can be directly weighted. In FL, the SVM models generated on different clients can be integrated by averaging the parameters and intercepts to meet the need of the server. The FedSVM algorithm is first proposed in \cite{bakopoulou2019federated} for mobile packet classification. We have adapted their FedSVM to our failure prediction problem. Its training process is shown in Fig. \ref{fig:FedSVM}. The algorithm of the client and server is given by Algo. \ref{algo:SVMClient} and Algo. \ref{algo:SVMServer}, respectively. The main steps of FedSVM are explained hereafter. 

\begin{itemize}
	\item[$\circ$] 
	\textbf{S1. Parameter initialization:} The server ($S$) notifies each client ($C_i$) to train models using the local data with random value of parameters and intercepts ($w_0$).
	\item[$\circ$] 
	\textbf{S2. Local training:} $C_i$ trains the local SVM model independently. 
	\item[$\circ$] 
	\textbf{S3. Return local SVM parameters:} $C_i$ returns locally trained parameters and intercepts ($w^i_t$) in this round to $S$.
	\item[$\circ$] 
	\textbf{S4. Compute global SVM parameters: } $S$ averages the trained values in two successive round $(\sum w^i_t + w_{t-1})/2$ as the value $w_t$ in the global model $\mathtt{M_G}$.
	\item[$\circ$] 
	\textbf{S5. Send global SVM parameter:}  $S$ sends the global value $w_t$ in the $\mathtt{M_G}$ to $C_i$ to train the model of next round.
	\item[$\circ$] 
	\textbf{S6. Repeat until convergence:} Repeat steps S2 - S5 after a specified number of iterations until convergence. The global model in the last round is sent to all clients as the training result.
\end{itemize}

\begin{algorithm}
	\caption{FedSVM - Client}
	\small
	\begin{algorithmic}[1]
		\Require Data set $D_i$ on client $i\ (i=1\ to\ k)$
		\Ensure Local SVM model on client $i$
		\State Initialize $w_0^i$ 
		\State $w_1^i \gets$ SVM($w_0^i$, $D_i$)  \Comment{S2}
		\State Send $w_1^i$ to server \Comment{S3}
		\For{$epoch\ t = 2\ to\ N$ } \Comment{S6}
		\State Receive global model $w_{t-1}$ \Comment{S5}
		\State $w_t^i\gets$  SVM($w_{t-1}$, $D_i$) \Comment{S2}
		\State Send $w_t^i$ to server \Comment{S3}
		\EndFor
	\end{algorithmic}
	\label{algo:SVMClient}
\end{algorithm}

\begin{algorithm}
	\caption{FedSVM - Server}
	\small
	\begin{algorithmic}[1]
		\Require Local SVM model
		\Ensure Global SVM model
		\State Inform all clients start training \Comment{S1}
		\State Receive $w_1^i$ from all clients 
		\State $w_1 \gets \sum_{i=1}^k \frac{w_{1}^{i}}{k}$ \Comment{S4}
		\State Send $w_1$ to all clients \Comment{S5}
		\For{$epoch\ t = 2\ to\ N$} \Comment{S6}
		\State Receive $w_t^i$ from all clients 
		\State $w_t \gets \frac{1}{2}(w_{t-1}+\sum_{i=1}^k \frac{w_t^i}{k})$ \Comment{S4}
		\State Send $w_t$ to all clients \Comment{S5}
		\EndFor
	\end{algorithmic}
	\label{algo:SVMServer}
\end{algorithm}

\subsection{Federated Random Forest}

The Federated Forest algorithm was first proposed by Liu et al. in 2020	 \cite{liu2020federated}. The random forest can be regarded as an integrated implementation of the decision tree. The core steps of the decision tree (calculating the purity of each feature, represented by Gini coefficients) only involve all data in a single feature and the classification label. Each client in VFL can do this on its local data. In our work, we have optimized the federated random forest algorithm in \cite{liu2020federated} to get better accuracy and efficiency by improving the steps of optimal feature selection and pruning. The improved method is illustrated hereafter.

\textbf{Optimal feature selection:} The optimal feature is the one with the smallest Gini coefficient according to the CART method in the decision tree. For continuous variables, it is necessary to try to divide all possible values when calculating the Gini coefficient. Here we reduce the calculation steps of this function and improve the efficiency of the algorithm.

\textbf{Pruning:} There is no explicit pruning process in \cite{liu2020federated}. Here we add a pre-pruning function, which enhances robustness and improves the efficiency of the overall algorithm. When the decision tree finds that the new node will not improve the accuracy, it will change this node to a leaf node.

\begin{figure*}[!htbp]
	\centering
	\includegraphics[height=0.35\textheight]{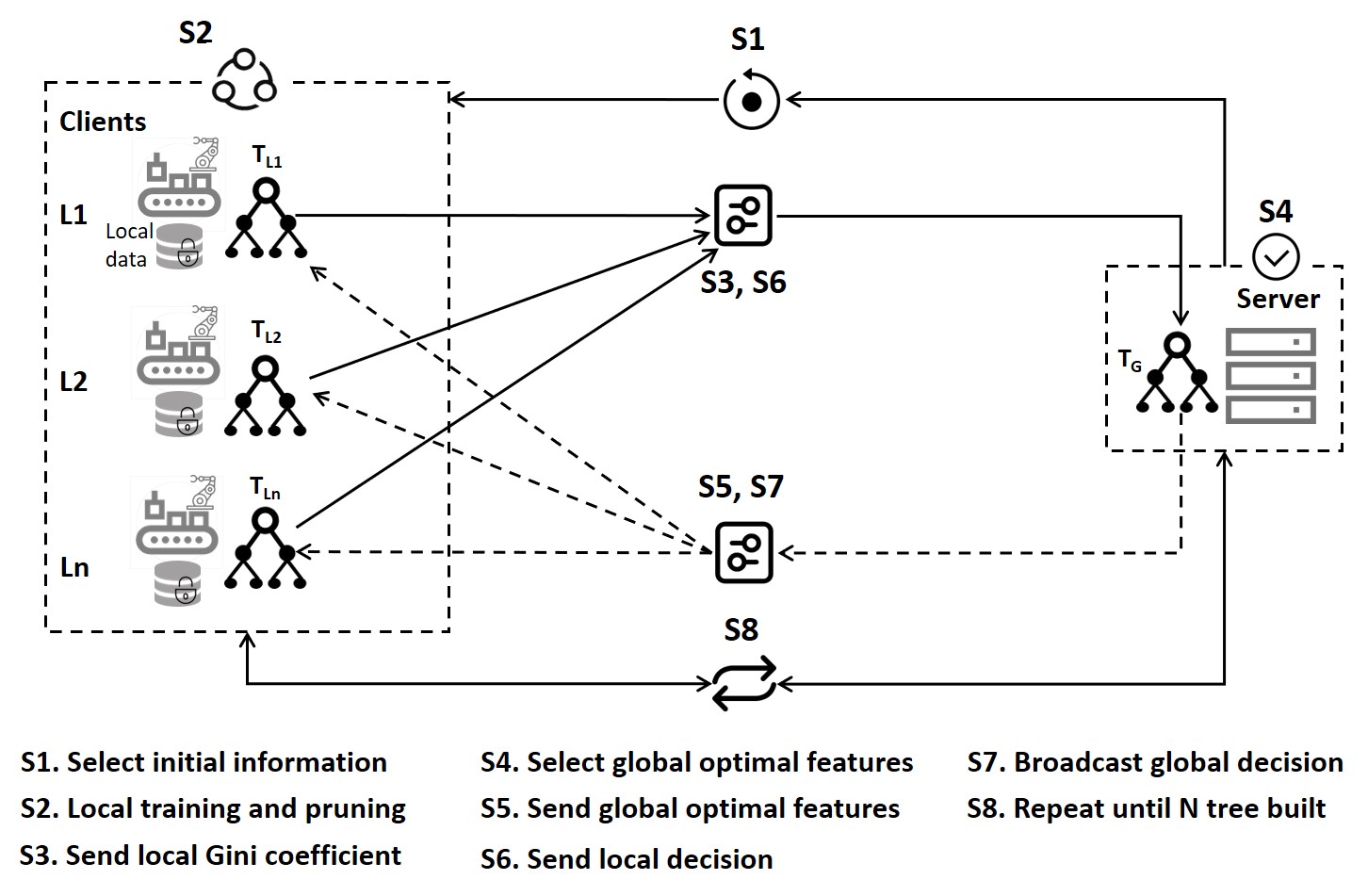}
	\caption{The Training Process of FedRF}
	\label{fig:FedRF}
\end{figure*}

The training process of our improved FedRF algorithm is given in Fig. \ref{fig:FedRF}. The algorithm of client and server is given by Algo. \ref{algo:RFClient} and Algo. \ref{algo:RFServer}, respectively. The main steps in its training process are explained hereafter.

\begin{itemize}
	\item[$\circ$] 
	\textbf{S1. Select initial information:} The server ($S$) randomly selects the feature subset $\mathtt{F'}$, the sample ID subset $\mathtt{D'}$ and test set $\mathtt{T}$, and notifies them to each client ($C_i$). 
	\item[$\circ$] 
	\textbf{S2. Local training and pruning:} $C_i$ knows which features have been selected, and does not know the number of features selected globally. $C_i$ calculates the Gini coefficient of each feature in $\mathtt{F'}$. Because the feature is only on the client, the client also knows the classification labels of all samples, so the Gini coefficient calculated by the client represents the global Gini coefficient. $C_i$ obtains the optimal division feature and corresponding Gini coefficient, and retain the division threshold of this feature. $C_i$ also checks whether the accuracy of the decision tree is improved on this optimal feature partition, and if there is no improvement, it judges that pre-pruning is needed.
	\item[$\circ$] 
	\textbf{S3. Upload local Gini coefficient:} $C_i$ uploads the Gini coefficient and pre-pruning information to $S$.
	\item[$\circ$] 
	\textbf{S4. Select global optimal features:} $S$ selects the globally optimal feature. 
	\item[$\circ$]
	\textbf{S5. Send optimal global features:} If pruning is not required, $S$ notifies the corresponding client to return the ID division result; otherwise, $S$ notifies all clients to perform pruning processing to form a leaf node.
	\item[$\circ$] 
	\textbf{S6. Send local decision:} When pruning is not required, the notified client returns the result of dividing the dataset with this feature (the ID set that falls into the left and right subtrees).
	\item[$\circ$] 
	\textbf{S7. Broadcast global decision:} $S$ notifies other clients of the ID division and forms the first node of the decision tree. $C_i$ knows the results of each ID division, but only knows the existence of local features in the global tree. $S$ knows the structure of the entire tree and the corresponding feature names of each node. $S$ and $C_i$ recursively establish new nodes and their left and right subtrees according to the current $\mathtt{F'}$ and $\mathtt{D'}$ until they form leaf nodes.
	\item[$\circ$] 
	\textbf{S8. Repeat until N tree built:} Iteratively build N decision trees to form a random forest.
\end{itemize}

\begin{algorithm}[!htbp]
	\caption{FedRF - Client}
	\begin{algorithmic}[1]
		\small
		\Require Data set $D$, feature set $F_{i}$ on client $i\ (i=1\ to\ k)$
		\Ensure Local FedRF model on $Ci$
		\For{$epoch\ t = 1\ to\ N$}
		\State Receive $F_{i}^{'} \subseteq F_{i}$, $D^{'} \subseteq D$, $T=D-D^{'}$ from $S$ 
		\Function{TreeBuild}{$D^{'}, F_{i}^{'}, T$}
		\If {all samples in $D^{'}$ have the same label}
		\State\Return $leaf\ node$
		\EndIf
		\State $BestF_{i}, GiniPara_{i}, threshold_{i} \gets $ Gini($D^{'}, F_{i}^{'}$) 
		\If {$BestF_{i}$ cannot improve the accuracy on $T$}
		\State Send $needPruning_{i}$ to $S$ \Comment{S2}
		\EndIf
		\State Send $BestF_{i}, GiniPara_{i}$ to $S$ \Comment{S3}
		\If {Receive $'Prune'$ message from $S$}
		\State\Return $leaf\ node$
		\ElsIf {Receive $'Success'$ message from $S$}
		\State Split $D^{'}$ with $BestF_{i}$, get $D_{l}^{'}, D_{r}^{'}, T_{l}, T_{r}$
		\State Send $D_{l}^{'}, D_{r}^{'}, T_{l}, T_{r}$ to $S$ \Comment{S6}
		\State leftTree  $\gets$  TreeBuild($D_{l}^{'}, F_{i}^{'}, T_{l}$)
		\State rightTree  $\gets$  TreeBuild($D_{r}^{'}, F_{i}^{'}, T_{r}$) \Comment{S7}
		\Else
		\State Receive $D_{l}^{'}, D_{r}^{'}, T_{l}, T_{r}$ from $S$
		\State leftTree  $\gets$  TreeBuild($D_{l}^{'}, F_{i}^{'}, T_{l}$)
		\State rightTree  $\gets$  TreeBuild($D_{r}^{'}, F_{i}^{'}, T_{r}$) \Comment{S7}
		\EndIf 
		\State\Return tree node
		\EndFunction
		\State Add this decision tree to forest 
		\EndFor \Comment{S8}
	\end{algorithmic}
	\label{algo:RFClient}
\end{algorithm}

\begin{algorithm}[!htbp]
	\caption{FedRF - Server}
	\begin{algorithmic}[1]
		\small
		\Require Sample ID, feature names on all clients
		\Ensure Global FedForest model
		\For{$epoch\ t = 1\ to\ N$}
		\For{$i=1\ to\ k$}
		\State Send $F_{i}^{'} \subseteq F_{i}$, $D^{'} \subseteq D$, $T=D-D^{'}$ to $C_i$ \Comment{S1}
		\EndFor 
		\Function{TreeBuild}{$D^{'}, T$}
		\If {all samples in $D^{'}$ have the same label}
		\State\Return $leaf\ node$
		\EndIf
		\State Receive $BestF_{i}^{k}, GiniPara_{i}^{k}, needPruning_{i}^{k}$  \Comment{S3}
		\State $BestF_{s} = Max(BestF_{i=1}^{k}, key = GiniPara_{i})$ \Comment{S4}
		\If {$needPruning_{s} == True$ }
		\State Send $'Prune'$ message to $C_i$
		\State\Return $leaf\ node$
		\EndIf
		\State Send $'Success'$ message to $C$ \Comment{S5}
		\State Receive $D_{l}^{'}, D_{r}^{'}, T_{l}, T_{r}$ from $C_i$
		\State Send $D_{l}^{'}, D_{r}^{'}, T_{l}, T_{r}$ to other $C_i$
		\State leftTree  $\gets$  TreeBuild($D_{l}^{'}, T_{l}$)
		\State rightTree  $\gets$  TreeBuild($D_{r}^{'}, T_{r}$)
		\State\Return tree node \Comment{S7}
		\EndFunction
		\State Add this decision tree to forest
		\EndFor \Comment{S8}
	\end{algorithmic}
	\label{algo:RFServer}
\end{algorithm}

The prediction process is described as follows. $C_i$ analyzes the new data ID according to the stored tree. For each sample, when an unknown node is encountered, the sample enters the left and right subtrees. When the encountered node is known (that means, the dividing feature of this node belongs to this client’s dataset), the sample enters the corresponding subtree according to the threshold of the feature. The final data that falls on the leaf node \emph{l} of the global tree \emph{t} is the intersection of all client data that falls on this leaf node.

We have experimented to evaluate the improvement of FedRF. 
In this experiment, we constructed a VFL scenario with two clients sharing data features using the Bosch dataset. There are 50 independent features in each client, which are extracted after performing the principal component analysis (PCA) from different production lines and the same 11154 samples. The experiment results are shown in Table \ref{tab:fedrf}. 
It can be seen that we get better prediction results than the federated random forest algorithm proposed in \cite{liu2020federated}. 
\begin{table}[!htbp]
	\centering
	\setlength{\extrarowheight}{2pt}
	\caption{Improvement of FedRF}    
	\scriptsize		
	\begin{tabular}{  p{1.4cm}  c  c  c  c  c }
		\toprule
		\textbf{Algo.} & \textbf{ACC} & \textbf{PRE} & \textbf{F1} & \textbf{MCC} & \textbf{AUC} \\ 
		\midrule
		FedForest \cite{liu2020federated} & 0.552 & 0.493 & 0.550 & 0.118 & 0.559 \\
		FedRF (Ours) & \fcolorbox{lightgray}{lightgray}{0.825} & \fcolorbox{lightgray}{lightgray}{0.802} & \fcolorbox{lightgray}{lightgray}{0.883} & \fcolorbox{lightgray}{lightgray}{0.593} & \fcolorbox{lightgray}{lightgray}{0.866} \\		
		\bottomrule
	\end{tabular}
	\label{tab:fedrf}
\end{table}


\section{Methodology}
\label{sec:methodo}

\subsection{Experiment Overview and Research questions}
\label{sec:rq}
\begin{figure*}[!htbp]
	\centering
	\includegraphics[height=0.27\textheight]{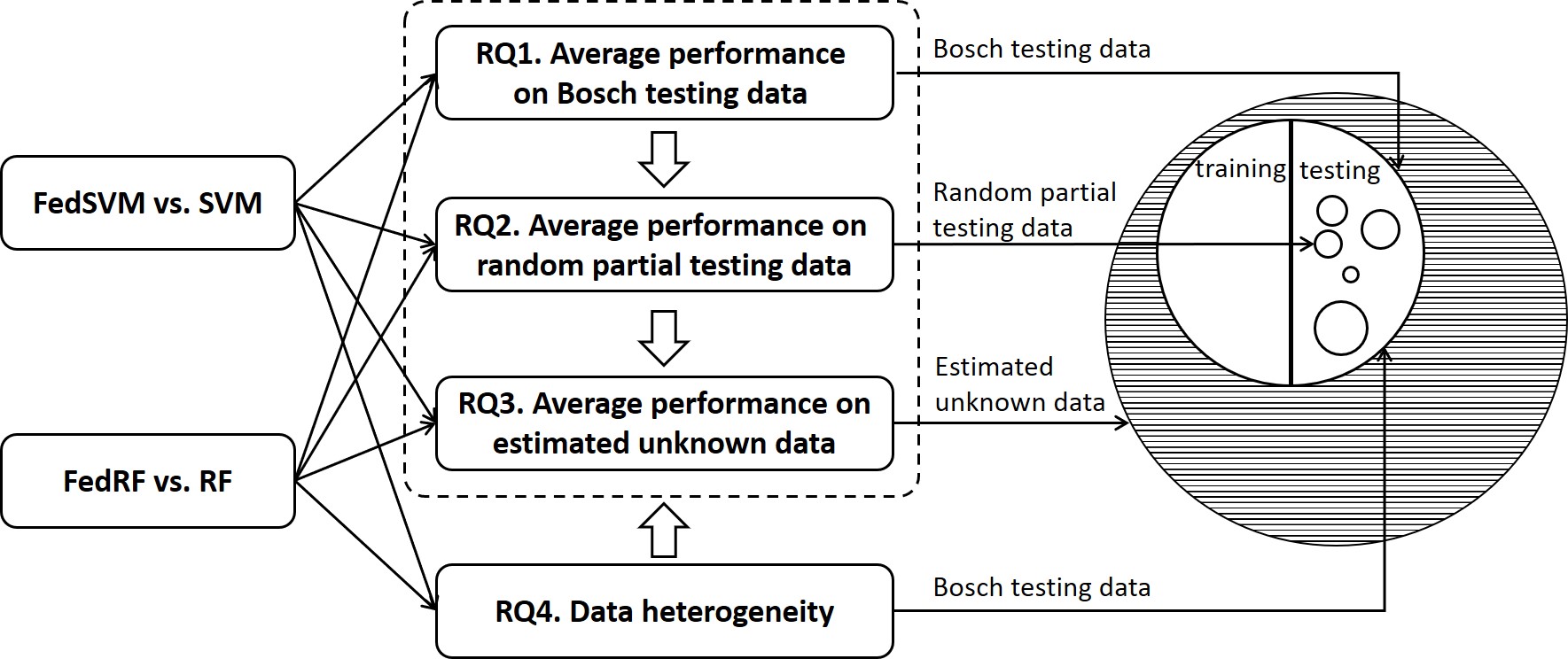} 
	\caption{Experiment Overview and Research Questions}
	\label{fig:methodo}
\end{figure*}

This paper studies whether there is a big difference between the effectiveness of FL and CL algorithms for failure prediction on the production line, and then whether FL algorithms can replace CL algorithms on this problem. To this end, we construct two production scenarios. One is HFL, where we compare FedSVM and SVM. The other is VFL, where we compare the effectiveness of FedRF and RF. We design four research questions (RQ), each of which concerns both HFL and VFL. The logical association between the four RQs is depicted, as shown in Fig. \ref{fig:methodo}.

\textbf{RQ1: On the whole Bosch testing dataset, is there a significant difference between FedSVM and SVM? We ask the same question for FedRF and RF.}

RQ1 is aimed at comparing the average performance of FL and CL algorithms on the whole Bosch testing dataset. If the difference between the value of each measurement is within the threshold $\delta$ ( $\delta = 0.1$), it can be considered that there is no significant difference between this pair of algorithms on the Bosch testing dataset.

\textbf{RQ2: On the random partial Bosch testing dataset, is there a significant difference between FedSVM and SVM? We ask the same question for FedRF and RF.}

RQ2 is aimed at comparing the average performance of the FL and CL algorithms on the random partial Bosch testing dataset that contains consecutive time-series samples. If the prediction results have no significant difference, it means that the pair of algorithms perform similarly on the random partial testing data.

\textbf{RQ3: On the estimated unknown Bosch data, is there a significant difference between FedSVM and SVM? We ask the same question for FedRF and RF.}

\begin{figure*}[!htbp]
	\centering
	\includegraphics[height=0.25\textheight]{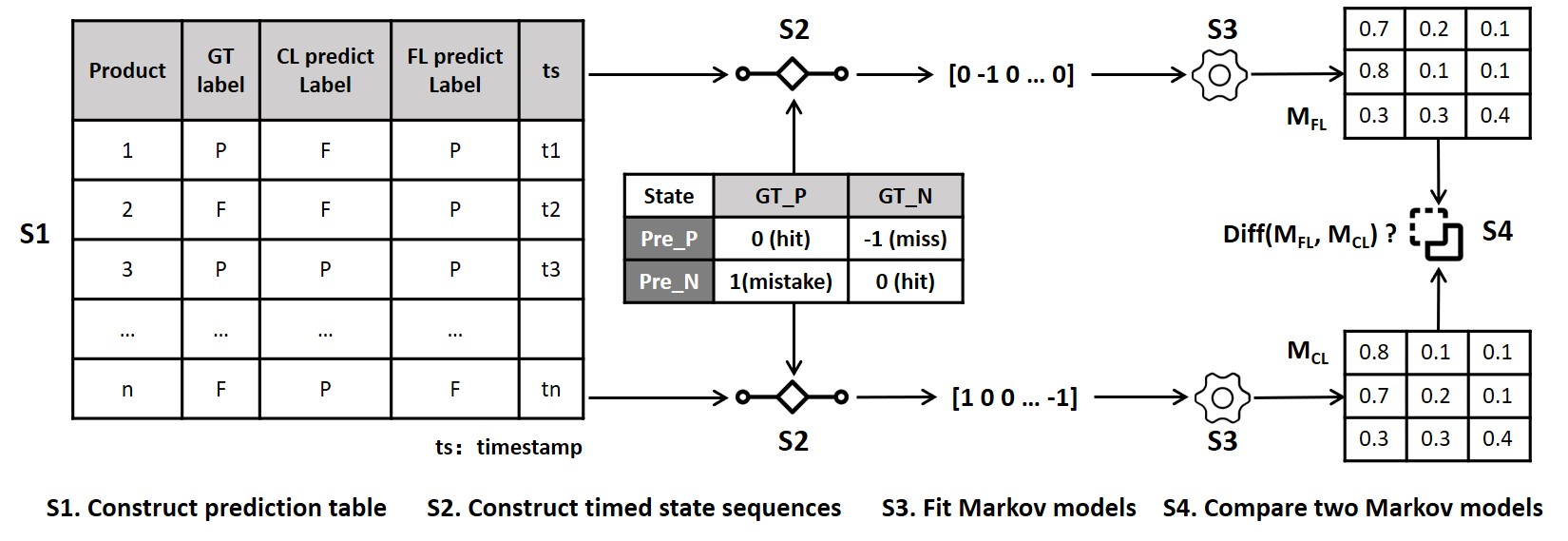}
	\caption{Comparing FL and CL algorithms on the Estimated Unknown Bosch Data}
	\label{fig:markov}
\end{figure*}

RQ3 is aimed at comparing the average performance of FL and CL algorithms on the estimated unknown Bosch data. To answer this RQ, we have designed a method to compare the error distribution of FL and CL algorithm on the estimated unknown data based on the given data, as shown in Fig. \ref{fig:markov}. This method consists of four main steps (S1-S4), explained hereafter.

\textbf{S1. Construct prediction table.} We first construct a prediction table containing the ground truth (GT) label, the prediction results (Pre) of CL and FL algorithms of products ordered by timestamp (ts). 

\textbf{S2. Construct timed state sequence.} The states represent the prediction error compared to the GT labels: 1) \emph{hit} state: GT positive + Pre positive or GT negative + Pre negative; 2) \emph{miss} state: GT positive + Pre negative; 3) \emph{mistake} state: GT negative + Pre positive. 

\textbf{S3. Fit Markov models.} We fit Markov model using the timed state sequence, and obtain $\mathtt{M_{FL}}$ and $\mathtt{M_{CL}}$, respectively. 

\textbf{S4. Compare FL and CL Markov models.} If the difference between the parameters of $\mathtt{M_{FL}}$ and $\mathtt{M_{CL}}$ is within the threshold $\delta$ ($\delta = 0.1$), the difference between FL and CL predictive models will maintain similar on estimated unknown Bosch data if the production line structure, manufacturing process, and quality control methods are not changed.

\textbf{RQ4: Is Bosch testing data heterogeneous? And what is the impact of data heterogeneity on the results of RQ1-RQ3?}

RQ4 analyzes the impact of data heterogeneity on the results of RQ1-RQ3. To evaluate the heterogeneity of data, we first obtain N groups of randomly selected continuous samples $\mathtt{D^i}$, and then uses the GT labels as states (GT positive state and GT negative state) to construct the state transition sequence $\mathtt{S^i_{GT}}$. Then we fit the Markov model $\mathtt{M^i_{GT}}$ using $\mathtt{S^i_{GT}}$, and perform DBSCAN clustering \cite{ester1996density} on the parameter matrix of $\mathtt{M^i_{GT}}$. If the clustering result shows multiple different clusters, it means that the data heterogeneity is strong. DBSCAN calculates the distance between two parameter matrices by converting the two matrices into a one-dimensional vector and calculating the angle between the two vectors.
If the included angle is less than the specified neighborhood distance threshold, the two-parameter matrices can be regarded as on the same cluster.

Based on the results of heterogeneity, the conclusions of RQ1-RQ3 (\emph{Y} means there is no significant difference between FL and CL, \emph{N} means there is significant difference between FL and CL) will be analyzed from the following four aspects.

\begin{itemize}
	\item[$\circ$] 
	If the data heterogeneity is strong, and the answers to RQ1-RQ3 are \emph{Y}, then the conclusion will be further strengthened. In our manufacturing scenario, the FL and CL algorithms can obtain similar prediction results and can replace each other.
	\item[$\circ$] 
	If the data heterogeneity is strong and some answer in RQ1-RQ3 is \emph{N}, it means that data heterogeneity may be one of the reasons for disturbing the  conclusions.
	\item[$\circ$] 
	If the data heterogeneity is weak and the answers to RQ1-RQ3 are \emph{Y}, it can be concluded that the FL algorithm can replace the CL one under the premise of data homogeneity. 
	\item[$\circ$] 
	If the data heterogeneity is weak and some answer in RQ1-RQ3 is \emph{N}, it means that the FL algorithm cannot replace CL even with homogeneous data. 
\end{itemize}

\subsection{Measurements}

MSA is one of the commonly used analysis methods in manufacturing \cite{montgomery2007introduction}. The learning-based failure prediction method can be seen as a product quality measurement system. Taking the Bosch company's product quality testing results as a benchmark, we can analyze the FL and CL algorithms through the MSA method. The evaluation measurements in MSA mainly include Accuracy (ACC), Precision (PRE), and Stability. On this basis, F1, AUC, and Matthew's Correlation Coefficient (MCC) are also involved. The six measurements we use in the empirical study are provided in Table \ref{tab:measurement}, where P represents the number of GT positive cases, N is the number of GT negative cases, TP (hit) is true positive, TN (correct rejection) is true negative, FP (false alarm) is false positive, and FN (miss) is false negative. 

\begin{table}[!htbp]
	\centering
	\setlength{\extrarowheight}{2pt}
	\caption{Measurements}    		
	\begin{tabular}{  c  p{5.3cm} }
		\toprule
		\textbf{Measurement} & \textbf{Formula} \\
		\midrule
		ACC  & \(\mathtt{\dfrac{TP + TN}{TP + FN + TN + FP}} \)   \\ [10pt]
		PRE  & \(\mathtt{ \dfrac{TP}{TP + FP}} \)  \\[10pt]
		F1 & \( \mathtt{\dfrac{2TP}{2TP + FP + FN}} \)  \\[10pt]
		MCC & \( \mathtt{\dfrac{TP \times TN - FP \times FN}{\sqrt{(TP + FP) (TP + FN) (TN + FP) (TN + FN)}}} \) \\[10pt]
		AUC & The area enclosed by the ROC curve and the reference line \\[10pt]
		Stability & The tendency of accuracy for N groups of random selected continuous data group. \\[10pt]
		\bottomrule
	\end{tabular}
	\label{tab:measurement}
\end{table}

MCC is suitable for evaluating the prediction results of unbalanced data. AUC is a commonly used two-category evaluation method, and its value is the area enclosed by the ROC curve and the reference line. The x-axis of a ROC curve is the false positive rate, and the y-axis of a ROC curve is the true positive rate. It shows the relationship between clinical sensitivity and specificity for every possible cut-off. It is a graph with: The x-axis showing 1-specificity \big($\mathtt{= FP/(FP+TN)}$\big) The y-axis showing sensitivity \big($\mathtt{= TP/(TP+FN)}$\big). The range of AUC value is between 0 and 1. 

Stability characterizes the ability of the measurement system to maintain a constant performance within a certain time range. We divide the test set into N groups (N = 10) on average according to the time-series and calculate the accuracy of these N groups to analyze the stability of the target algorithm. 

\subsection{Bosch Dataset}

The Bosch dataset is one of the largest public manufacturing datasets (14.3Gb) on Kaggle. The Bosch dataset consists of a training set (1184687 samples) and a test set (1183748 samples). Each sample has three types of features: categorical feature, numeric feature, and date feature, and a two-category classification label indicating the faulty status of a product. 

The date feature gives the time stamp of each site through which the product passes. Based on the date feature, it is possible to analyze the timed behavior. Existing studies have constructed LSTM neural network models based on time-series information \cite{huang2019enhancing,moldovan2019time,liu2020adversarial}. Since we focus on empirical research on FL without time information, the date features are excluded from the scope of our study. Nevertheless, the time-series information can still provide time-continuity data for empirical research on the estimated unknown data.

The data features are shown in Table \ref{tab:bosch}. There are 1184687 products in the dataset, including 1177808 positive samples (Pos) and 6879 negative ones (Neg). In the feature space, the dataset has 968 features. These features are collected from 51 workstations (S1-S51) on 4 production lines L0-L3 that contain 168, 513, 42, and 245 features, respectively. L1\_S24\_F1695 indicates that the feature No.1695 was observed at the No.24 workstation of the L1 production line. The label Response represents whether the final product is qualified (label = 0) or not (label = 1). From the perspective of data items, only a few items (0.58\%) in the dataset result are qualified, and the data is extremely unbalanced. The data is also extremely sparse, with only a few workstations collecting data from most products \cite{carbery2019new}.

\begin{table}[!htbp]
	\centering
	\setlength{\extrarowheight}{2pt}
	\caption{Bosch dataset}    		
	\begin{tabular}{  c  c | c  c  c  c }
		\toprule
		\multicolumn{2}{c |}{\textbf{Sample Num (1184687)}} & \multicolumn{4}{c}{\textbf{Features Num (968)}}  \\ 
		\textbf{Positive} & \textbf{Negative} & \textbf{L0} & \textbf{L1} & \textbf{L2} & \textbf{L3}\\
		\midrule
		1177808 & 6879 &  168 & 513 & 42 & 245 \\
		\bottomrule
	\end{tabular}
	\label{tab:bosch}
\end{table}

\subsection{Experiment Scenarios of HFL and VFL}

To compare FedSVM and SVM on the testing data, we form four workshops A, B, C, and D that do not share manufacturing data but have the same structure and configuration of the production line. Under the premise that the product features are aligned, they contain different samples, and each sample has a unique ID.  Each workshop can be regarded as an independent client. Table \ref{tab:RQ1-SVM_data} shows the data used in this experiment. The number of samples on the 4 clients is basically the same. In the experiment, the data set is randomly distributed to all clients on average, each of which possesses 3000 positive and 1500 negative samples. The four workshops have 713 common features of data. 

\begin{table}[!htbp]
	\centering
	\setlength{\extrarowheight}{2pt}
	\caption{Data Description of HFL Scenario}    		
	\begin{tabular}{  c  c  c  c  }
		\toprule
		\textbf{Sample Num.} & \textbf{Positive} & \textbf{Negative} & \textbf{Feature Num.}  \\ 
		\midrule
		\textbf{Total} & 12000 & 6000  & \multirow{2}{2cm}{713 selected}  \\
		\textbf{Per client} & 3000 & 1500 &  \\
		\bottomrule
	\end{tabular}
	\label{tab:RQ1-SVM_data}
\end{table}

To compare FedRF and RF on the given testing data, it is assumed that the first three production lines (L0, L1, L2) of the Bosch dataset belong to the same organization O1, and the L3 production line belongs to another organization O2. O1 and O2 are independent. Since the samples in O1 and O2 have the same ID and different characteristics, VFL is applied. The data set used in the FedRF experiment is shown in Table \ref{tab:RQ1-RF_data}. We extract the features belonging to L0, L1, and L2 as a group, and the features belonging to L3 as another group. Then we perform principal component analysis (PCA) to reduce the feature dimension in O1/O2 from 723/245 to 22/22, respectively. According to the overall analysis of PCA results, the first 22 dimensions can represent more than 95\% of the variance, so reducing to 22 dimensions, respectively, which will not cause too much information loss \cite{zhang2016predict}. The positive and negative sample numbers are 12000 and 6000, respectively.

\begin{table}[!htbp]
	\centering
	\setlength{\extrarowheight}{2pt}
	\caption{Data Description for VFL Scenario}    	
	\begin{tabular}{ cccl }
		\toprule
		\textbf{Sample Num} & \textbf{Positive} & \textbf{Negative} & \textbf{Feature Num} \\ 
		\midrule
		\textbf{Total} & \multirow{3}{1cm}{12000} & \multirow{3}{1cm}{6000}   & 44 (F1 - F44) \\
		\textbf{Client O1} & & & 22 (F1 - F22)\\
		\textbf{Client O2} &  & & 22 (F23 - F44) \\
		\bottomrule
	\end{tabular}
	\label{tab:RQ1-RF_data}
\end{table}


\section{Experiment Results}
\label{sec:exp}

Our experiments investigate and answer four research questions. 

\subsection{RQ1: Comparison on Bosch Testing Data}

\subsubsection{FedSVM vs. SVM}

\begin{figure}[!htbp]
	\centering
	\includegraphics[height=0.28\textheight]{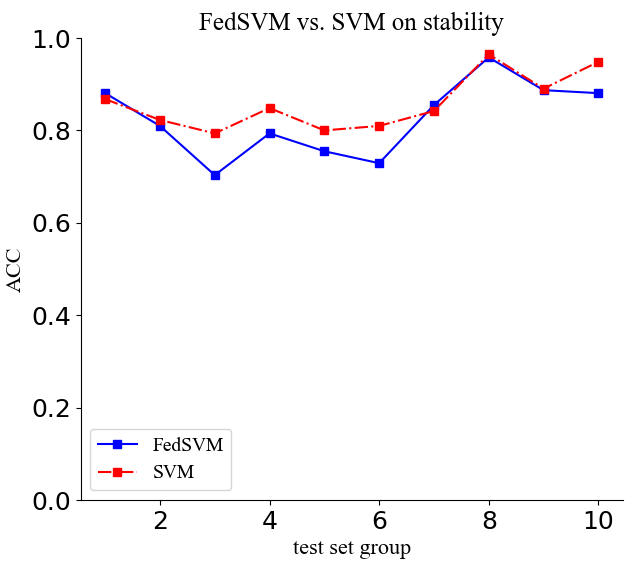}
	\caption{FedSVM vs. SVM on Stability}
	\label{fig:R1_SVM_STAB}
\end{figure}

Table \ref{tab:RQ1-SVM_res} shows the experimental results. To ensure that the data set and the algorithm kernel are consistent, the baseline is the result of FedSVM when the number of clients is set to 1, ie., FedSVM degenerates into SVM. Under this premise, the results of FedSVM and SVM are compared. When analyzing the stability, the experiment data is ordered by time-series and then divided into 10 groups on average. The measurements of each group are calculated. The overall results of stability are shown in Fig. \ref{fig:R1_SVM_STAB}. The average value of stability is above 0.7 for FedSVM and SVM. The variance of the stability is given as Stability (Var) in Table \ref{tab:RQ1-SVM_res}. 

\begin{table}[!htbp]
	\centering
	\setlength{\extrarowheight}{2pt}
	\caption{Experiment Results of RQ1 on FedSVM/SVM}
	\scriptsize   	
	\begin{tabular}{ c c c c c c p{0.8cm} }
		\toprule
		\textbf{Algo.} & \textbf{ACC} & \textbf{PRE} & \textbf{F1} &\textbf{MCC} & \textbf{AUC} & \textbf{Stab. (Var)}  \\ 
		\midrule
		\texttt{FedSVM} & 0.825 & 0.859 & 0.869 & 0.607 & 0.807 & 0.005  \\
		\texttt{SVM}    & 0.859	& 0.828	& 0.903	& 0.690	& 0.902	& 0.003 \\
		\texttt{Diff}   & -0.034& 0.031&-0.046	&-0.083	&-0.095	& 0.002 \\
		\bottomrule
	\end{tabular}
	\label{tab:RQ1-SVM_res}
\end{table}

The value difference of ACC, Precision, F1, MCC, AUC, and Stability (Var) between FedSVM and SVM are all within a threshold value of 0.1, respectively. Therefore, the conclusion of RQ1 on FedSVM and SVM is that \emph{the prediction results of FedSVM and SVM on the Bosch testing data are not significantly different}. 

In Figure \ref{fig:R1_SVM_STAB}, FedSVM performs slightly better than SVM on two test groups, namely 1 and 7. Similar results are also shown for the Precision measure in Table \ref{tab:RQ1-SVM_res}. This effect is only shown in the HFL scenario, where the clients contribute local data features to the central server. If the amount of data on each client is small, FL's effect can be better than CL because FL expands the number of IID data samples in the HFL scenario.

\subsubsection{FedRF vs. RF}

\begin{figure}[!htbp]
	\centering
	\includegraphics[height=0.28\textheight]{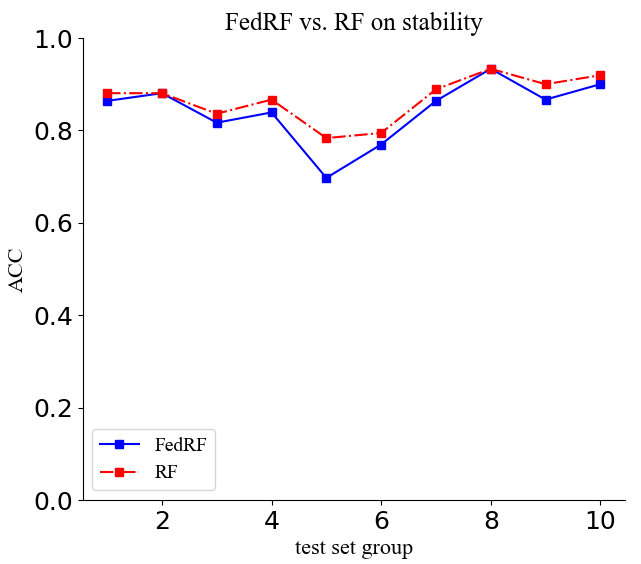}
	\caption{FedRF vs. RF on Stability}
	\label{fig:R1_RF_STAB}
\end{figure}

The experimental results are shown in Table \ref{tab:RQ1-RF_res}. FedRF degenerates into RF by combining two clients so that the effect of FedRF and RF can be compared under the premise of ensuring the consistency of the testing data and the algorithm kernel. When analyzing the stability, the test set is divided into 10 groups on average according to the time-series, and the measurements of each group are calculated. The variance of the stability is given in Table \ref{tab:RQ1-RF_res}, and the overall results of stability are shown in Fig. \ref{fig:R1_RF_STAB}.

\begin{table}[!htbp]
	\centering
	\setlength{\extrarowheight}{2pt}
	\caption{Experiment Results of RQ1 on FedRF/RF}    
	\scriptsize	
	\begin{tabular}{ c c c c c c p{0.8cm} }
		\toprule
		\textbf{Algo.} & \textbf{ACC} & \textbf{PRE} & \textbf{F1} &\textbf{MCC} & \textbf{AUC} & \textbf{Stab. (Var)}  \\ 
		\midrule
		\textbf{FedRF} & 0.843 & 0.808 & 0.894 & 0.659 & 0.902 & 0.004 \\
		\textbf{RF} & 0.868 &	0.836 &	0.909 &	0.713 &	0.912 &	0.002 \\
		\textbf{Diff} &-0.025 &-0.028 &-0.015 &-0.054 &-0.010 &0.002 \\
		\bottomrule
	\end{tabular}
	\label{tab:RQ1-RF_res}
\end{table}

The value difference of ACC, Precision, F1, MCC, AUC, and Stability (Var) on each group are all within 0.1. Therefore, \emph{the prediction results of FedRF and RF on the Bosch testing data are not significantly different.}

\subsection{RQ2: Comparison on Random Partial Bosch Testing Data}
In order to compare the average difference between the FL and CL algorithms in randomly sampling partial data, we randomly select N groups (N=100) of testing samples. Each group contains L time-continuous samples. The starting point S of the group is randomly set. L is randomly generated from the interval [300, 1000]. 

\subsubsection{FedSVM vs. SVM}

\begin{figure}[!htbp]
	\centering
	\subfloat[ACC Values of FedSVM and SVM]{
		\label{subfig:svm1}
		\includegraphics[width=0.47\textwidth]{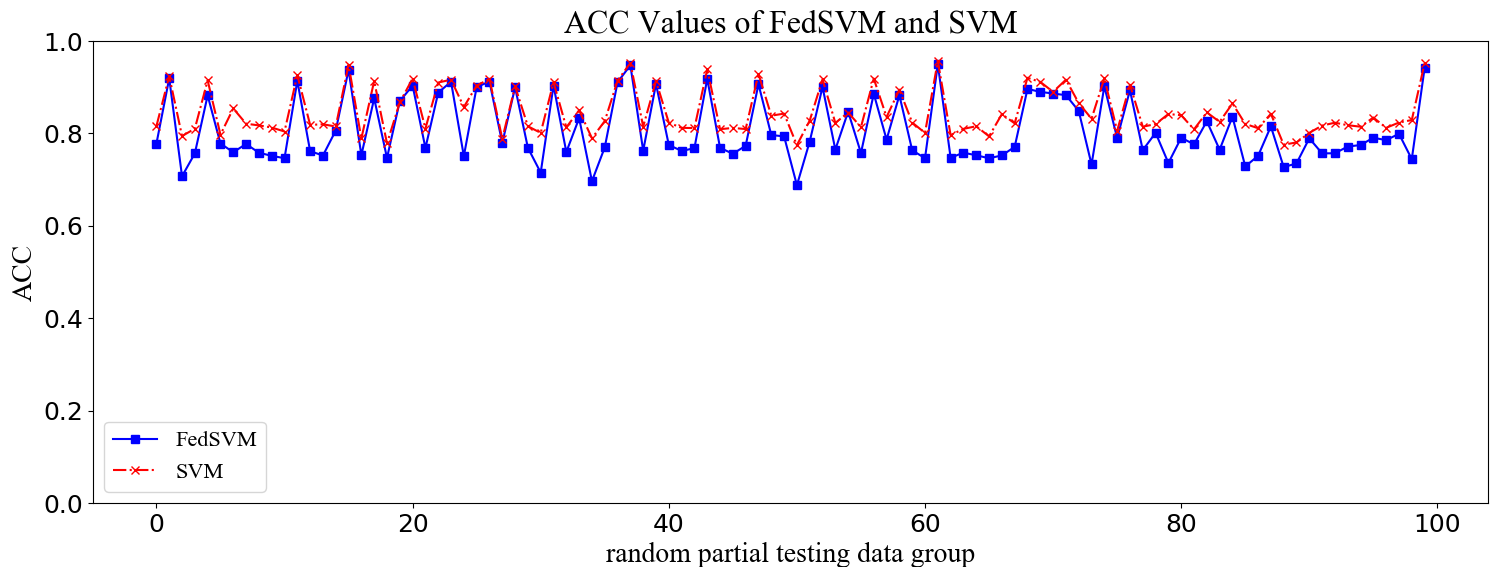} } 
	
	\subfloat[PRE Values of FedSVM and SVM]{
		\label{subfig:svm2}
		\includegraphics[width=0.47\textwidth]{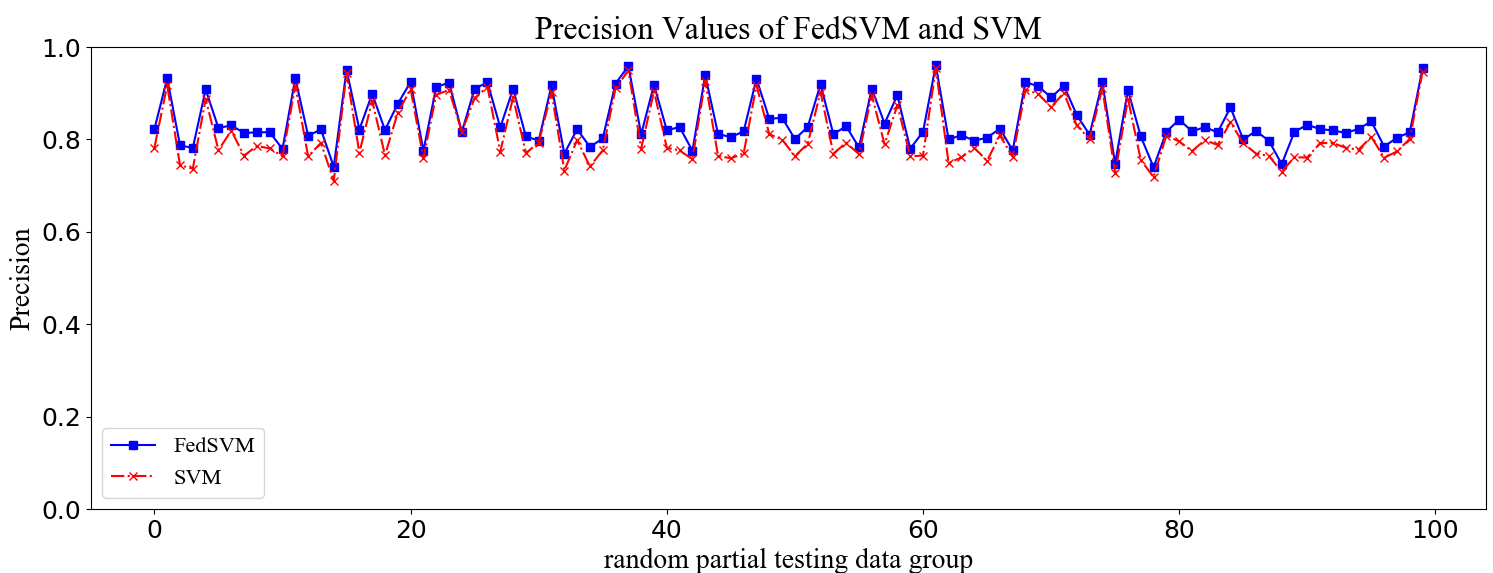} } 
	
	\subfloat[F1 Values of FedSVM and SVM]{
		\label{subfig:svm3}
		\includegraphics[width=0.47\textwidth]{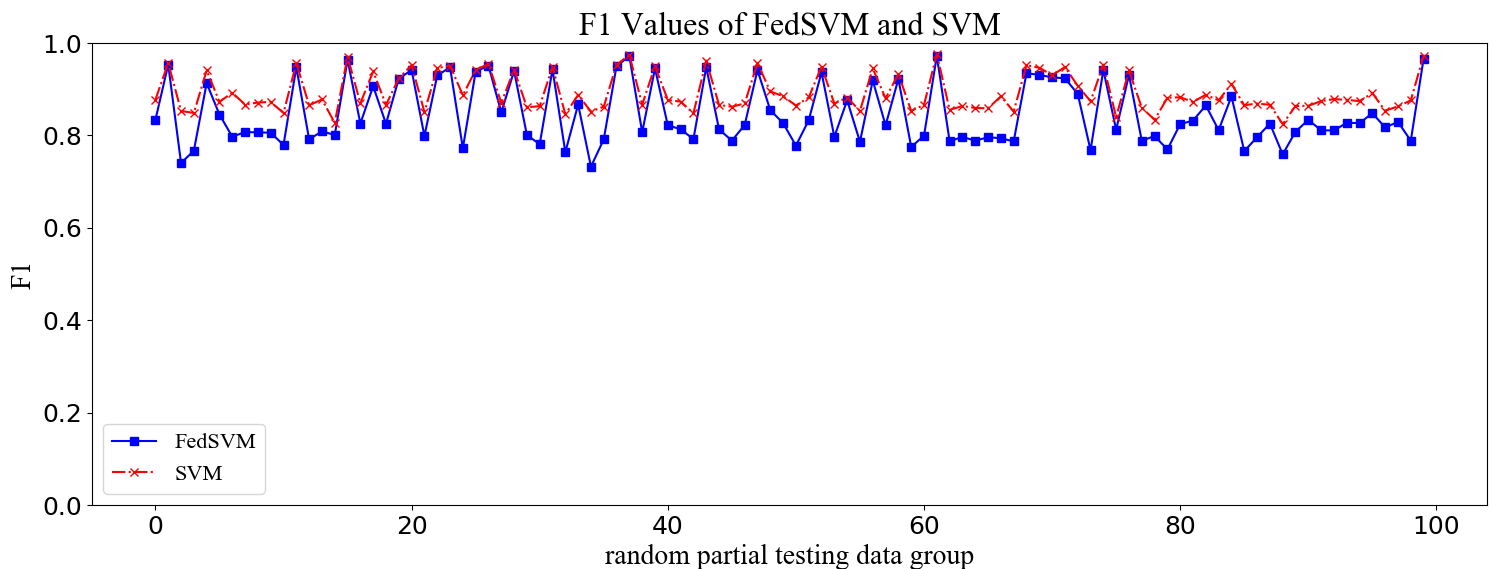}}
	
	\subfloat[MCC Values of FedSVM and SVM]{
		\label{subfig:svm4}
		\includegraphics[width=0.47\textwidth]{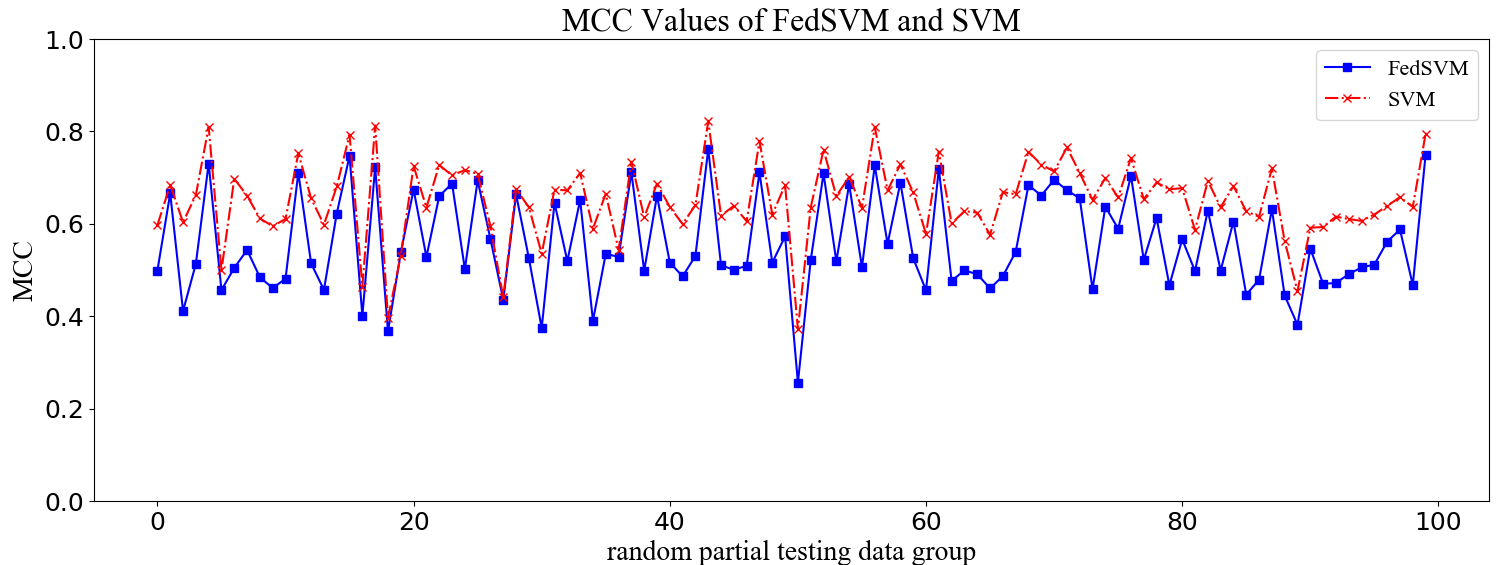}}
	
	\subfloat[AUC Values of FedSVM and SVM]{
		\label{subfig:svm5}
		\includegraphics[width=0.47\textwidth]{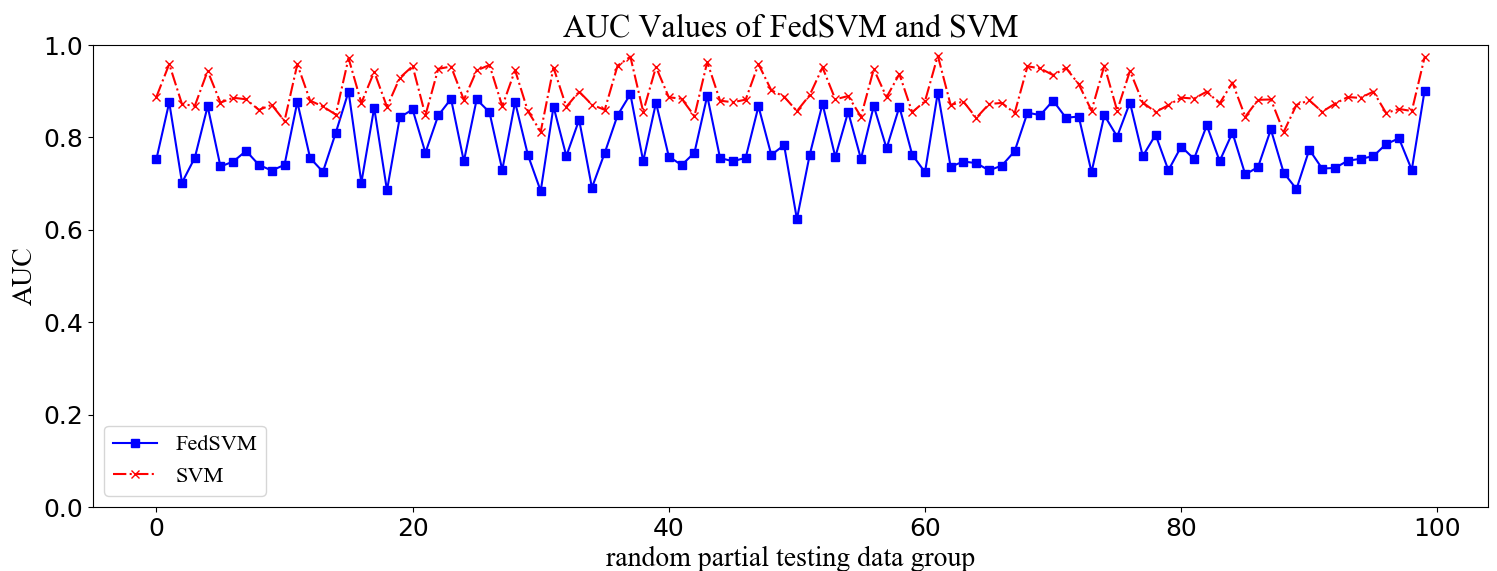}}
	\caption{FedSVM vs. SVM on Random Partial Data}
	\label{fig:R2_SVM}
\end{figure}

The line charts of the prediction results of FedSVM and SVM on 100 groups are shown in Fig. \ref{fig:R2_SVM}. In all charts, the abscissa is the serial number of the 100 groups. The ordinate is the value of the measurement on each group. The solid blue line and red dot-dash line represent the prediction result of FedSVM and SVM, respectively. Through the charts, we can see the range and difference of the FedSVM and SVM on the prediction results of each group.

\begin{figure}[!htbp]
	\centering
	\subfloat[ACC Difference]{
		\label{subfig:svm1diff}
		\includegraphics[width=0.22\textwidth]{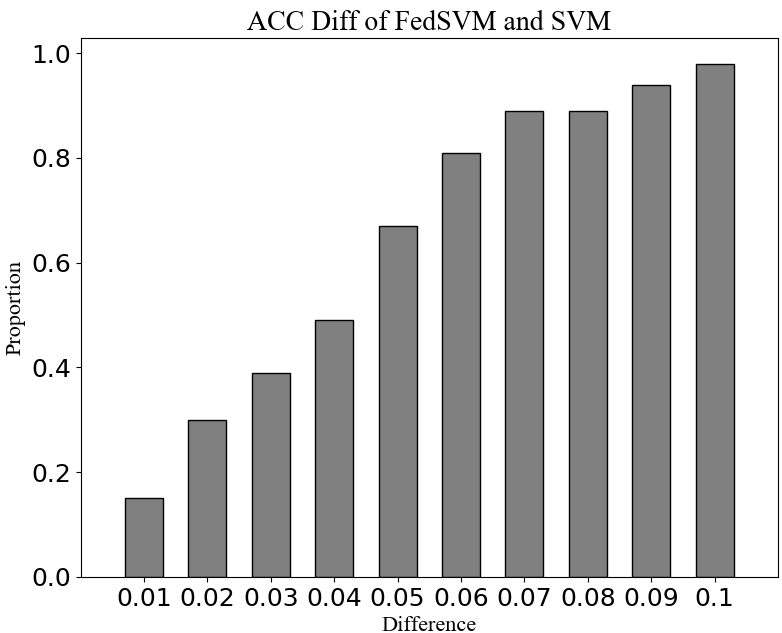} } 
	\subfloat[PRE Difference]{
		\label{subfig:svm2diff}
		\includegraphics[width=0.22\textwidth]{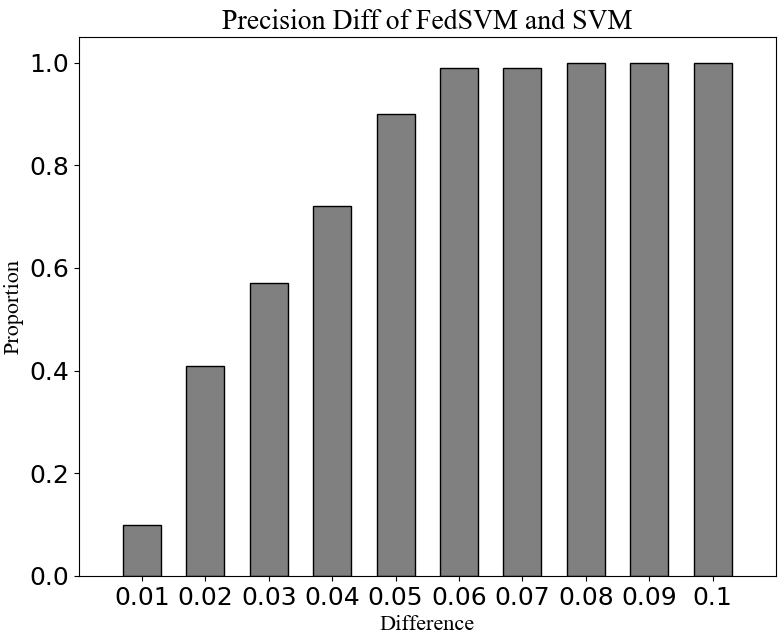} } 
	
	\subfloat[F1 Difference]{
		\label{subfig:svm3diff}
		\includegraphics[width=0.22\textwidth]{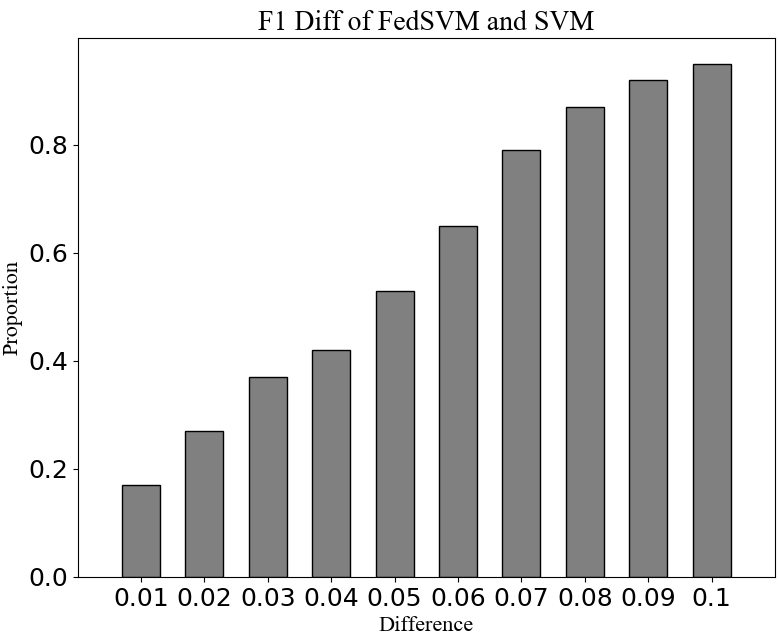}}
	\subfloat[MCC Difference]{
		\label{subfig:svm4diff}
		\includegraphics[width=0.22\textwidth]{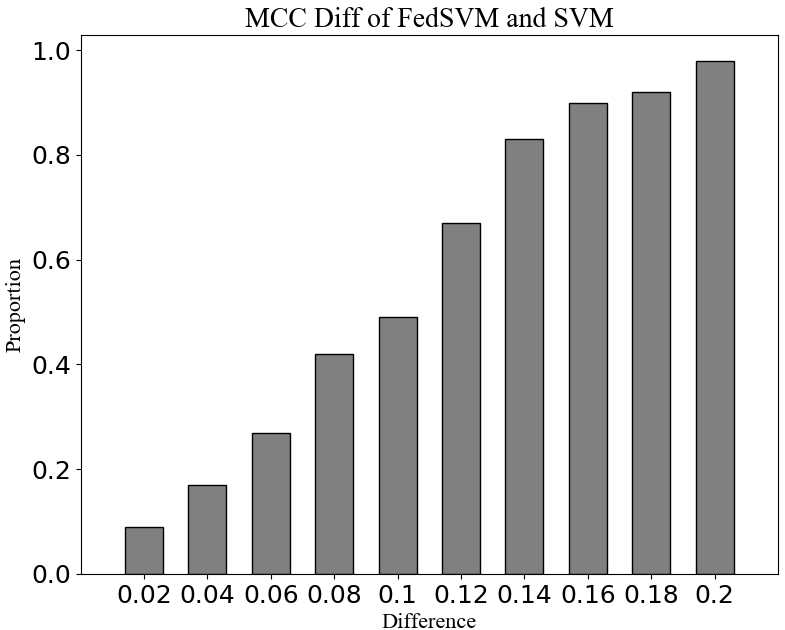}}
	
	\subfloat[AUC Difference]{
		\label{subfig:svm5diff}
		\includegraphics[width=0.22\textwidth]{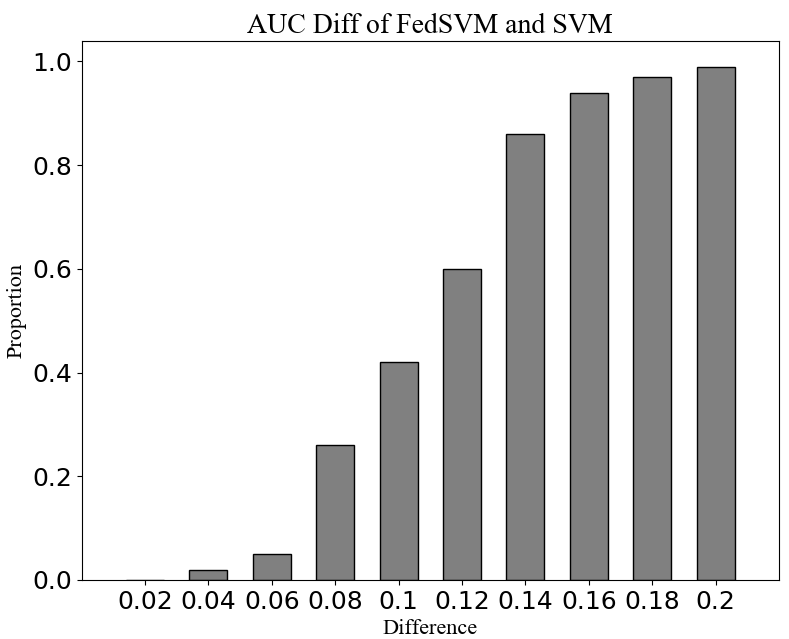}}
	\caption{Difference of FedSVM and SVM on Random Partial Testing Data}
	\label{fig:R2_SVM_diff}
\end{figure}

We draw a histogram to visualize the average difference between FedSVM and SVM on each measurement for 100 groups of data, as shown in Fig. \ref{fig:R2_SVM_diff}. Based on all results, the difference values in ACC, PRE, and F1 are all within 0.1. The difference values in MCC and AUC are mostly within 0.2. It is thus validated that \emph{the performance of FedSVM and SVM on random partial Bosch testing data is not significantly different}. 

\subsubsection{FedRF vs. RF}

\begin{figure}[!htbp]
	\centering
	\subfloat[ACC Values of FedRF and RF]{
		\label{subfig:ff1}
		\includegraphics[width=0.47\textwidth]{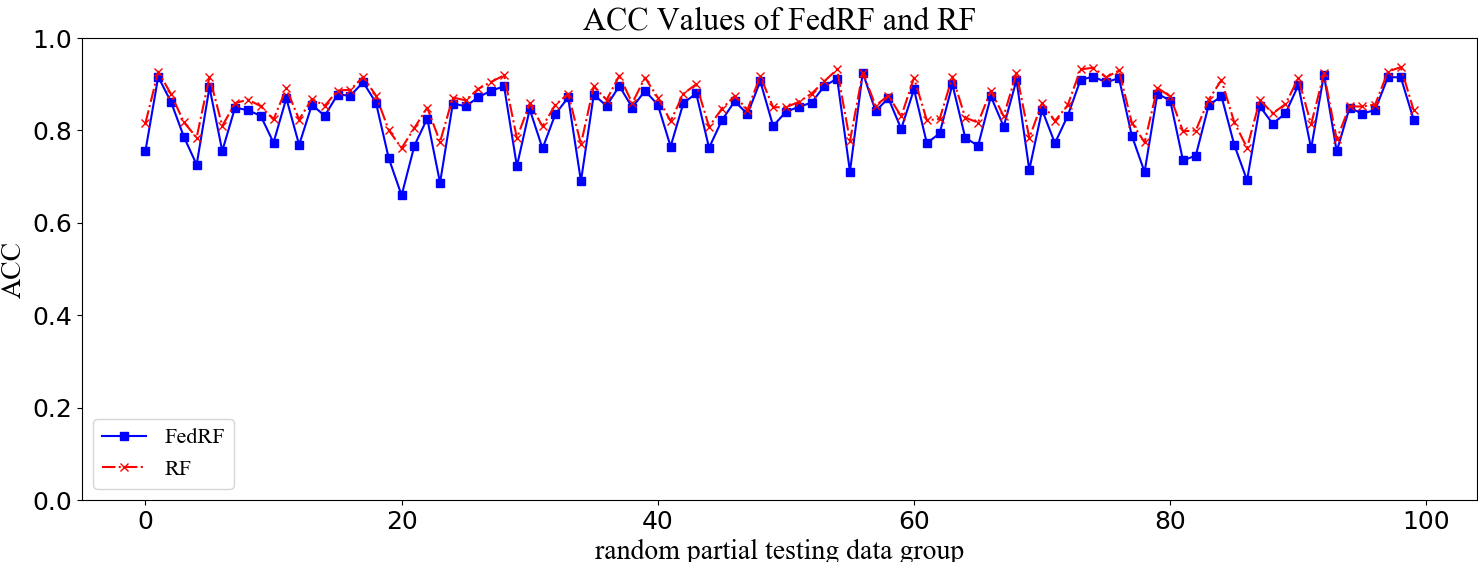} } 
	
	\subfloat[PRE Values of FedRF and RF]{
		\label{subfig:ff2}
		\includegraphics[width=0.47\textwidth]{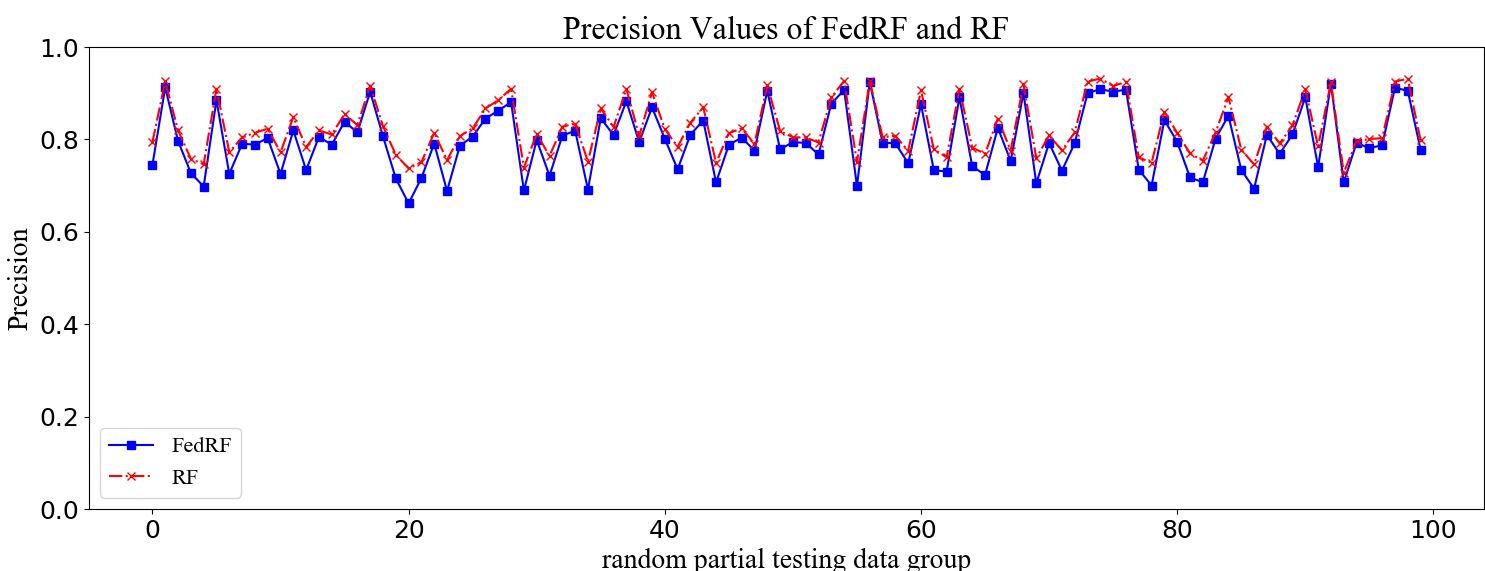} } 
	
	\subfloat[F1 Values of FedRF and RF]{
		\label{subfig:ff3}
		\includegraphics[width=0.47\textwidth]{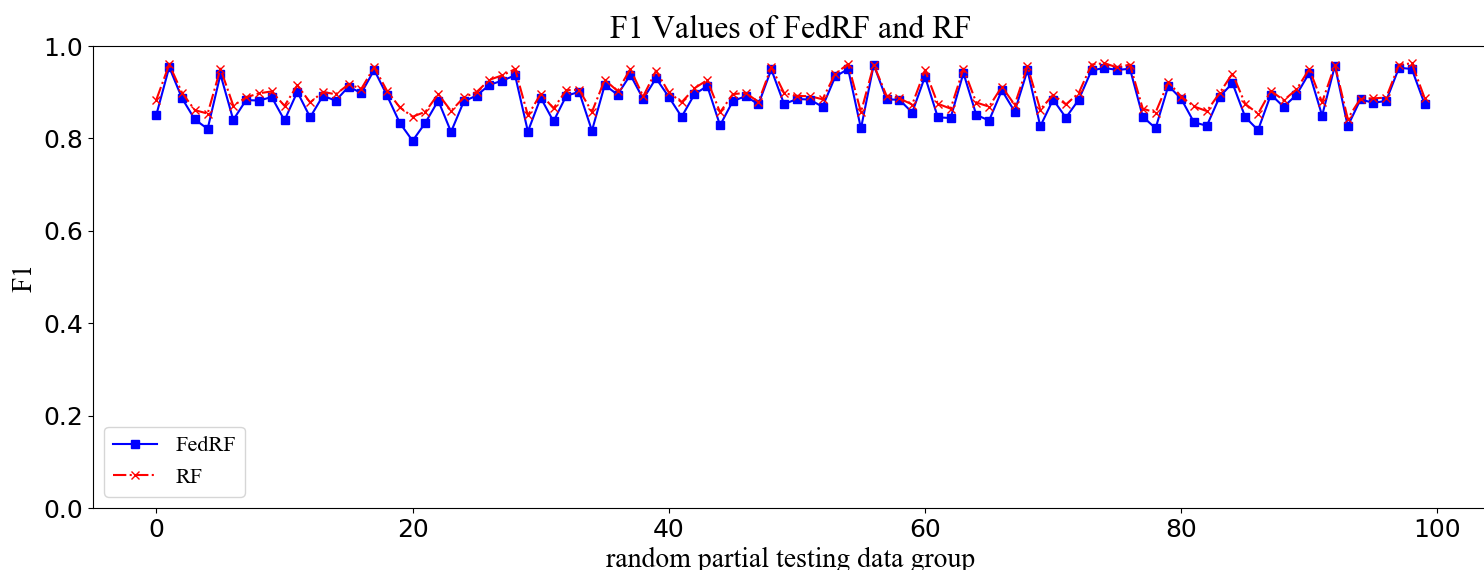}}
	
	\subfloat[MCC Values of FedRF and RF]{
		\label{subfig:ff4}
		\includegraphics[width=0.47\textwidth]{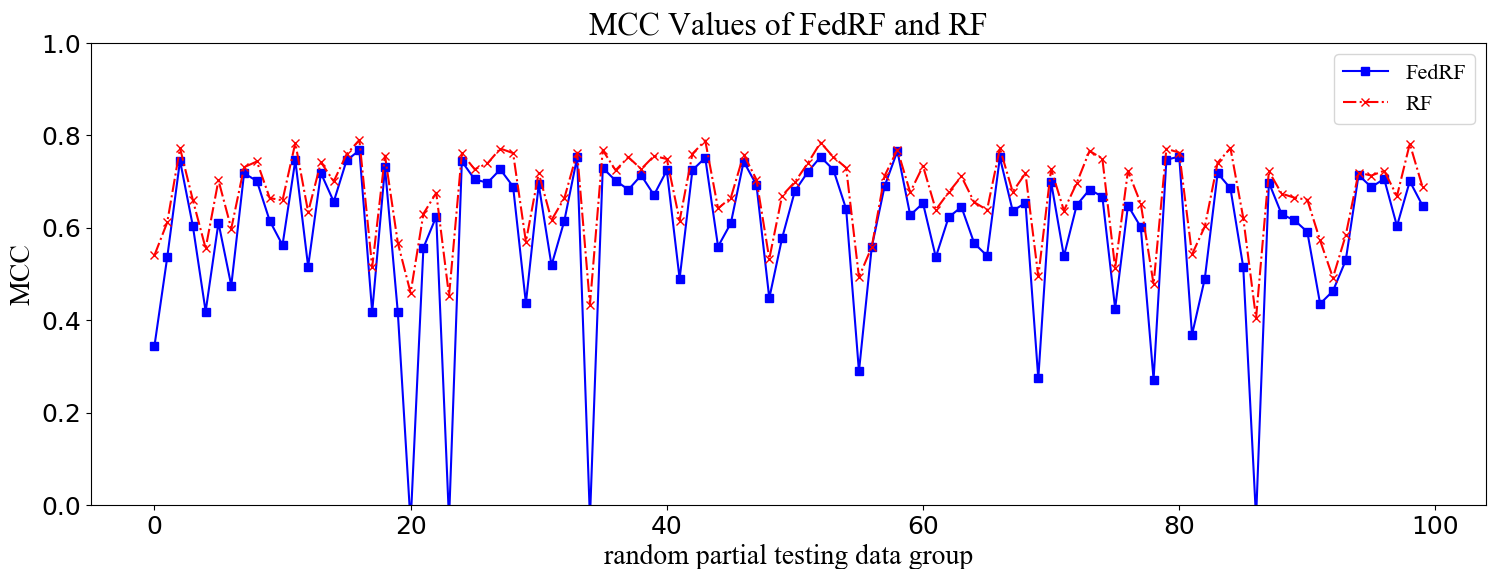}}
	
	\subfloat[AUC Values of FedRF and RF]{
		\label{subfig:ff5}
		\includegraphics[width=0.47\textwidth]{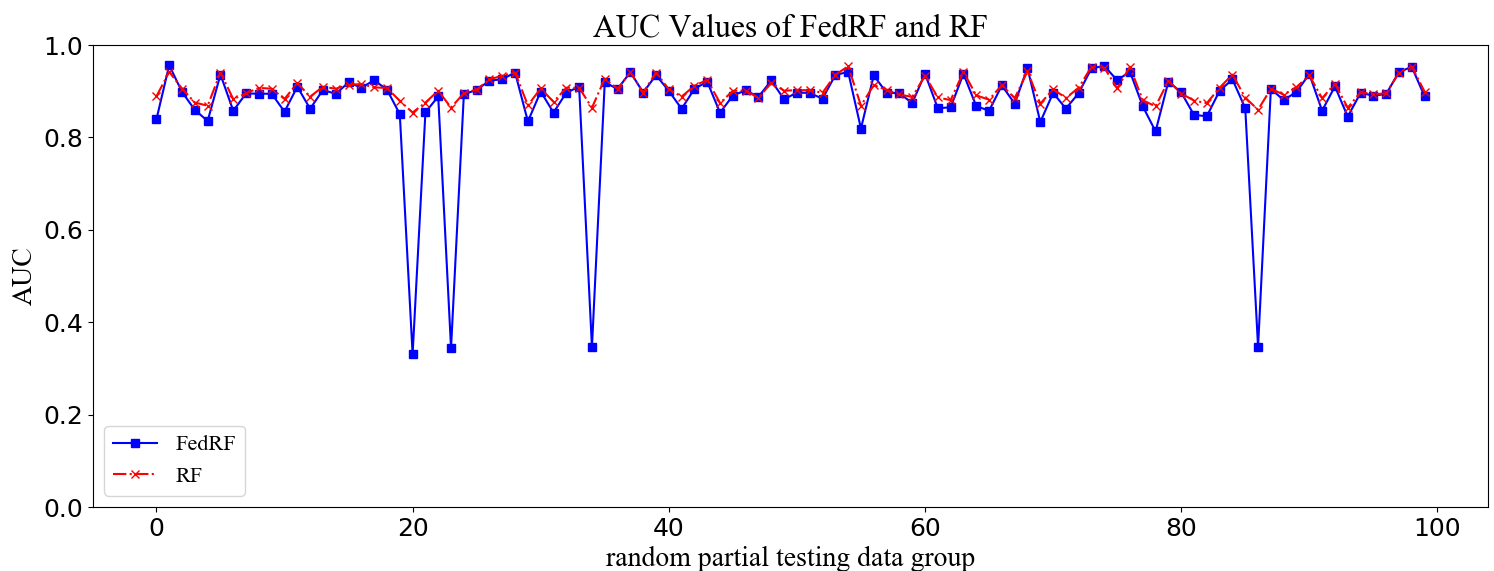}}
	\caption{FedRF vs. RF on Random Partial Data}
	\label{fig:R2_RF}
\end{figure}

The line charts of FedRF and RF on 100 groups of samples are shown in Fig. \ref{fig:R2_RF}. In all line charts, the abscissa is the number of groups. The ordinate is the value of the measurements on each group. The solid blue line and red dot-dash line represent the prediction result of the FedRF and RF, respectively.

\begin{figure}[!htbp]
	\centering
	\subfloat[ACC Difference]{
		\label{subfig:rf1diff}
		\includegraphics[width=0.22\textwidth]{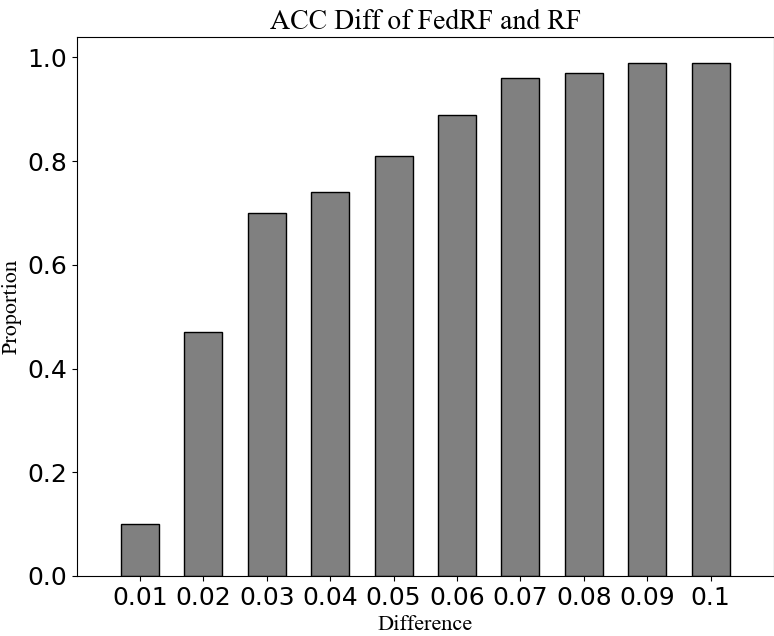} } 
	\subfloat[PRE Difference]{
		\label{subfig:rf2diff}
		\includegraphics[width=0.22\textwidth]{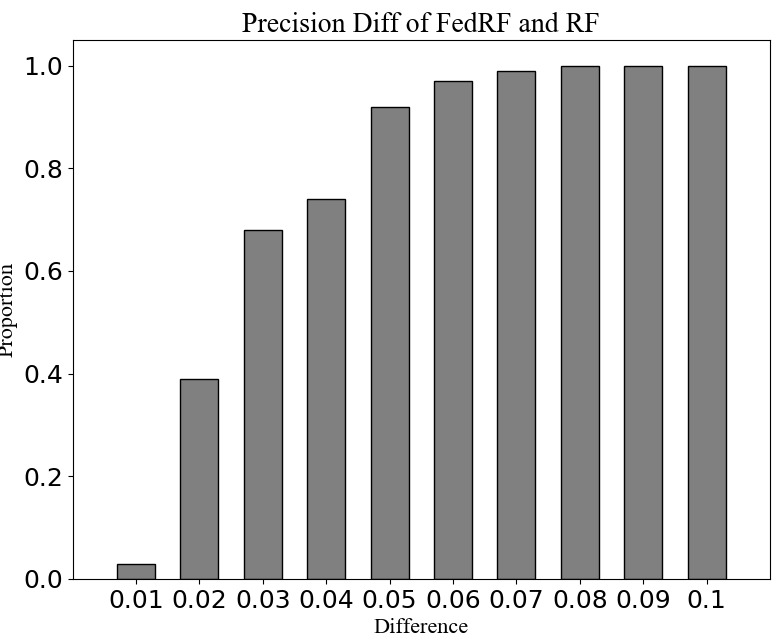} } 
	
	\subfloat[F1 Difference]{
		\label{subfig:rf3diff}
		\includegraphics[width=0.22\textwidth]{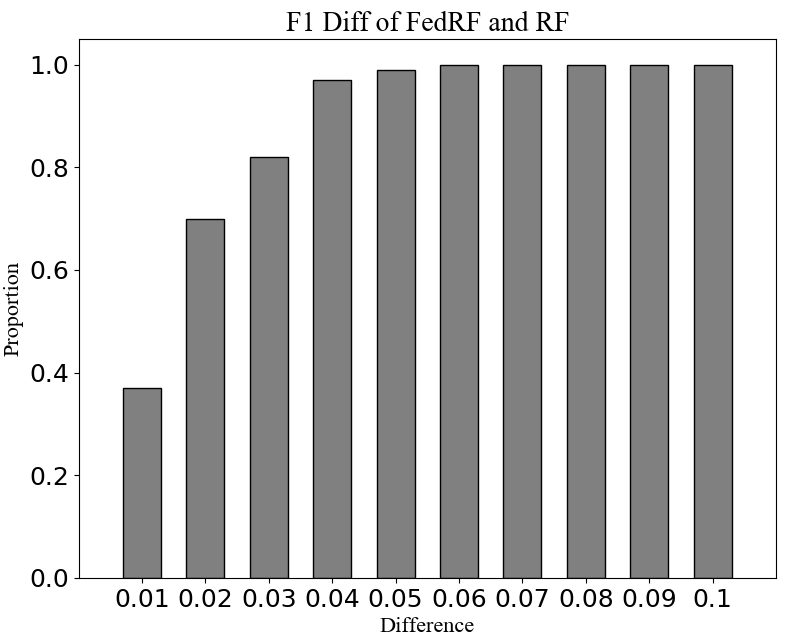}}
	\subfloat[MCC Difference]{
		\label{subfig:rf4diff}
		\includegraphics[width=0.22\textwidth]{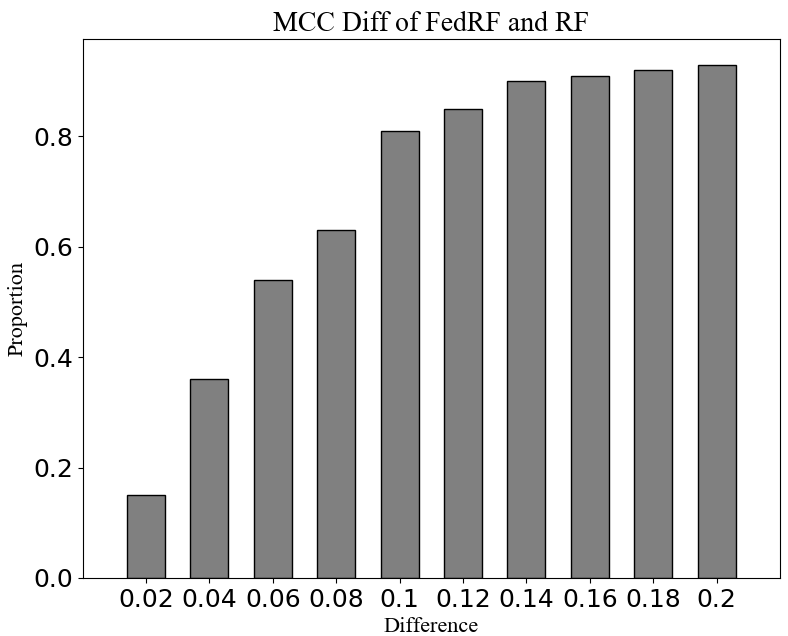}}
	
	\subfloat[AUC Difference]{
		\label{subfig:rf5diff}
		\includegraphics[width=0.22\textwidth]{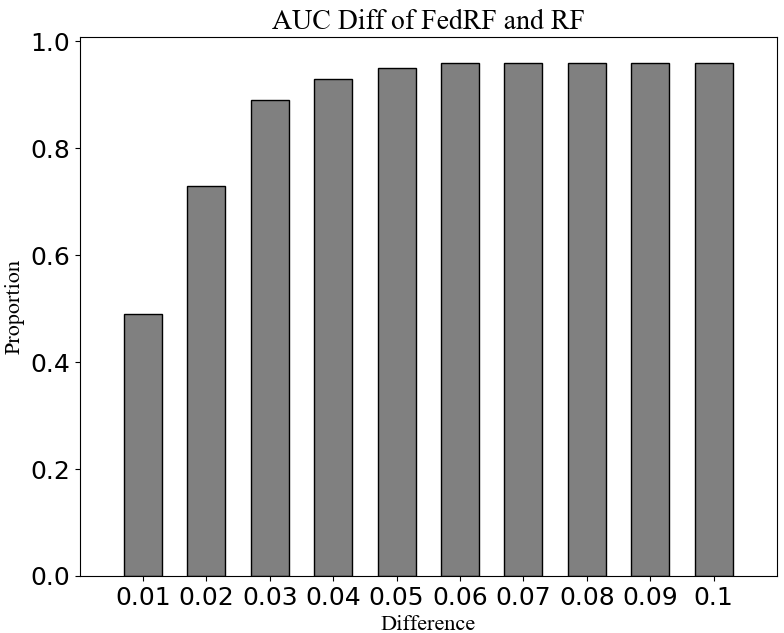}}
	\caption{Difference of FedRF and RF on Random Partial Testing Data}
	\label{fig:R2_RF_diff}
\end{figure}

We draw a histogram of the differences between FedRF and RF on each measurement for 100 groups of data, as shown in Fig. \ref{fig:R2_RF_diff}. Based on all results, the difference values in ACC, PRE, F1, and AUC are all within 0.1. The difference values in MCC are within 0.2, and the MCC values of over 80\% groups are within 0.1. It is thus validated that \emph{the performance of FedRF and RF on the random partial Bosch testing data is not significantly different.}

\subsection{RQ3: Comparison on Estimated Unknown Bosch Data}

This experiment is used to analyze whether the FL and CL algorithms can maintain no significant difference on the estimated unknown Bosch data. This experiment is conducted following the method presented in Sect. \ref{sec:methodo}.

\subsubsection{FedSVM vs. SVM}

We first constructed a timed state sequence indicating the prediction error using FedSVM and SVM, respectively, based on which their Markov models are fitted. The parameters of two Markov models are given in Table \ref{tab:RQ3-SVM_resa} and \ref{tab:RQ3-SVM_resa}. The value difference between each pair of parameters is calculated in Table \ref{tab:RQ3-SVM_resc}.
The average difference value between each pair of parameters is 0.054. The maximum difference value is 0.096. It means that there exists no significant difference between the two Markov models, and the prediction error between FedSVM and SVM are almost equally distributed on the estimated unknown Bosch data. Therefore, \emph{the performance of FedSVM and SVM in the subsequent Bosch production was not significantly different.}

\begin{table}[!htbp]
	\centering
	\setlength{\extrarowheight}{2pt}
	\caption{Experiment Results of RQ3 on FedSVM/SVM }   
	\subfloat[Markov Model Parameters of Prediction Error using FedSVM]{ 	
		\begin{tabular}{ p{1.5cm} p{1.5cm} p{1.5cm} p{1.5cm} }
			\toprule
			\textbf{State} & \textbf{Hit} & \textbf{Miss} & \textbf{Mistake}   \\ 
			\midrule
			\textbf{Hit} & 0.882 & 0.113 & 0.005  \\
			\textbf{Miss} & 0.715 & 0.280  & 0.005 \\
			\textbf{Mistake} & 0.813 & 0.062 & 0.125 \\
			\bottomrule
			\label{tab:RQ3-SVM_resa}
	\end{tabular}}\\[-4ex]
	
	\subfloat[Markov Model Parameters of Prediction Error using SVM ]{ 
		\begin{tabular}{ p{1.5cm} p{1.5cm} p{1.5cm} p{1.5cm} }
			\toprule
			& \textbf{Hit} & \textbf{Miss} & \textbf{Mistake}   \\ 
			\midrule
			\textbf{Hit} & 0.851 & 0.077 & 0.072  \\
			\textbf{Miss} & 0.690 & 0.225 & 0.085  \\
			\textbf{Mistake} & 0.717 & 0.125 & 0.158  \\
			\bottomrule
			\label{tab:RQ3-SVM_resb}
	\end{tabular}}\\[-4ex]
	
	\subfloat[Comparison of Two Markov Model Parameters]{ 
		\begin{tabular}{ p{1.5cm} p{1.5cm} p{1.5cm} p{1.5cm} }
			\toprule
			& \textbf{Hit} & \textbf{Miss} & \textbf{Mistake}   \\ 
			\midrule
			\textbf{Hit}     & 0.031 & 0.036 & 0.067  \\
			\textbf{Miss}    & 0.025 & 0.055 & 0.080  \\
			\textbf{Mistake} & 0.096 & 0.063 & 0.033 \\
			\bottomrule
			\label{tab:RQ3-SVM_resc}	
	\end{tabular}}\\[-4ex]
	\label{tab:RQ3-SVM_res}
\end{table}

\subsubsection{FedRF vs. RF}

We first constructed a timed state sequence indicating the prediction error using FedRF and RF, respectively, based on which their Markov models are fitted. The parameters of two Markov models are given in Table \ref{tab:RQ3-RF_resa} and \ref{tab:RQ3-RF_resa}. The value difference between each pair of parameters is calculated in Table \ref{tab:RQ3-RF_resc}.
The average difference value between each pair of parameters is 0.035. The maximum difference value is 0.100. It means that there exists no significant difference between the two Markov models, and the prediction error between FedRF and RF are equally distributed on the estimated unknown Bosch data. Therefore, \emph{the performance of FedRF and RF in the subsequent Bosch production was not significantly different.}

\begin{table}[!htbp]
	\centering
	\setlength{\extrarowheight}{2pt}
	\caption{Experiment Results of RQ3 on FedRF/RF }    
	\subfloat[Markov Model Parameters of Prediction Error using FedRF]{ 	
		\begin{tabular}{ p{1.5cm} p{1.5cm} p{1.5cm} p{1.5cm} }
			\toprule
			& \textbf{Hit} & \textbf{Miss} & \textbf{Mistake}   \\ 
			\midrule
			\textbf{Hit} & 0.869 & 0.130 & 0.001  \\
			\textbf{Miss} & 0.701 & 0.299 & 0.000  \\
			\textbf{Mistake} & 1.000 & 0.000 & 0.000  \\
			\bottomrule
			\label{tab:RQ3-RF_resa}			
	\end{tabular}}\\[-4ex]
	
	\subfloat[Markov Model Parameters of Prediction Error using RF]{ 	
		\begin{tabular}{ p{1.5cm} p{1.5cm} p{1.5cm} p{1.5cm} }
			\toprule
			& \textbf{Hit} & \textbf{Miss} & \textbf{Mistake}   \\ 
			\midrule
			\textbf{Hit} & 0.888 & 0.109 & 0.003  \\
			\textbf{Miss} & 0.735 & 0.263 & 0.002  \\
			\textbf{Mistake}   & 0.900 & 0.100 & 0.000  \\
			\bottomrule
			\label{tab:RQ3-RF_resb}			
	\end{tabular}}\\[-4ex]
	
	\subfloat[Comparison of Two Markov Model Parameters]{ 	
		\begin{tabular}{p{1.5cm} p{1.5cm} p{1.5cm} p{1.5cm} }
			\toprule
			& \textbf{Hit} & \textbf{Miss} & \textbf{Mistake}   \\ 
			\midrule
			\textbf{Hit} & 0.019 & 0.021 & 0.002  \\
			\textbf{Miss} & 0.034 & 0.036& 0.002  \\
			\textbf{Mistake} & 0.100 & 0.100 & 0.000   \\
			\bottomrule
			\label{tab:RQ3-RF_resc}			
	\end{tabular}}\\[-4ex]
	\label{tab:RQ3-RF_res}
\end{table}

\subsection{RQ4: Heterogeneity of Bosch Testing Data}

This experiment is used to evaluate the heterogeneity of the testing data used in the experiments of RQ1-RQ3. We follow the method proposed in Sect. \ref{sec:methodo} to experiment. One hundred groups of time-ordered consecutive samples were randomly selected. According to the GT label of the sample, the qualified sample is set to state S0, and the unqualified sample is set to state S1, forming a state sequence. For the state sequence of each group of samples, the Markov model of k=1 step and k=2 steps are established, respectively. The DBSCAN algorithm is used to cluster and analyze all the 1-step and 2-step parameter matrices. If the clustering results show multiple different clusters, it can explain that there is heterogeneity within the testing data. 

\subsubsection{Heterogeneity of the Testing Data used in FedSVM}

The experimental results are shown in Table \ref{tab:RQ4-SVM_res}. There are outliers in the test results, so the sum of the samples in each cluster is less than 100. It can be seen that the testing data is divided into 2 clusters. The difference between clusters is no less than the distance threshold, which means that the data used in the experiment of FedSVM is heterogeneous. It enforces the conclusion that \emph{FedSVM and SVM have no significant difference on heterogeneous manufacturing data for the problem of failure prediction}.

\begin{table}[!htbp]
	\centering
	\setlength{\extrarowheight}{2pt}
	\caption{Experiment Results of RQ4 on FedSVM}   
	\scriptsize 	
	\begin{tabular}{ c c p{1.8cm} p{1.3cm} p{1.3cm} }
		\toprule
		\textbf{Step} & \textbf{Clusters} & \textbf{Distance threshold ($^{\circ}$)} & \textbf{Contour factor} & \textbf{Samples / cluster} \\ 
		\midrule
		k = 1 & 2 & \centering 5 &  0.456 & 92 / 6	\\
		k = 2	& 2 & \centering 8 &  0.421 & 89 / 6 \\
		\bottomrule
	\end{tabular}
	\label{tab:RQ4-SVM_res}
\end{table}

According to the experiment results and the statement of RQ4 in Section \ref{sec:rq}, the data heterogeneity in the VFL scenario is relatively strong and the answer to RQ1-RQ3 are positive. This means in our VFL manufacturing scenarios, the FL and CL algorithms can obtain similar prediction results and can thus replace each other under the premise of data heterogeneity. 

\subsubsection{Heterogeneity of the Testing Data used in FedRF}

The experimental results are shown in Table \ref{tab:RQ4-RF_res}. It can be seen that the testing data is divided into 2 clusters. The difference between clusters is no less than the distance threshold, which means that the data used in the experiment of FedRF is heterogeneous. It enforces the conclusion that \emph{FedRF and RF have no significant difference on heterogeneous manufacturing data for the problem of failure prediction}.

\begin{table}[!htbp]
	\centering
	\setlength{\extrarowheight}{2pt}
	\caption{Experiment Results of RQ4 on FedRF}    
	\scriptsize 			
	\begin{tabular}{ c c p{1.8cm} p{1.3cm} p{1.3cm} }
		\toprule
		\textbf{Step} & \textbf{Clusters} & \textbf{Distance threshold ($^{\circ}$)} & \textbf{Contour factor} & \textbf{Samples / cluster} \\ 
		\midrule
		k = 1	& 2 & \centering 4 & 0.524 & 94 / 6	\\
		k = 2	& 2	& \centering 8 & 0.473	& 78 / 22 \\
		\bottomrule
	\end{tabular}
	\label{tab:RQ4-RF_res}
\end{table}

According to the experiment results and the statement of RQ4 in Section \ref{sec:rq}, the data heterogeneity in the HFL scenario is relatively strong and the answer to RQ1-RQ3 are positive. This means in our HFL manufacturing scenarios, the FL and CL algorithms can obtain similar prediction results and can thus replace each other under the premise of data heterogeneity. 


\section{Threats to Validity}
\label{sec:threats}

Following common guidelines for empirical studies \cite{runeson2009guidelines,yin2017case}, we discuss threats to the validity of our study.

Threats to internal validity are mainly concerned with uncontrolled factors. These factors may impact the results and reduce their creditability. In this work, the main threat to internal validity is potential defects in the implementation of our own algorithm and the reimplementation of other algorithms (mainly the federated SVM and federated random forest algorithms). To reduce this threat, we reimplement the baseline techniques by following their papers. We carefully review all our code and experiment scripts by several developers to ensure their correctness. However, there is always a small possibility of defects, which introduces risk to the result’s correctness.

Threats to external validity are mainly concerned with whether the performance of our techniques can still hold in other experimental settings. In this work, the main threat to external validity is the impact of data preprocessing method. As we mentioned in Sect. \ref{sec:intro}, the existing works revealed that the data preprocessing on the Bosch dataset has a significant impact on the prediction result. In our work, we have applied PCA to find the principle features. For example, in the VFL scenario, according to the PCA analysis, only 22 features are kept in each client. But, these features represent more than 95\% of the variance of data. Another threat to external validity is the experimental methodology we designed. In order to answer whether the FL algorithm can replace the CL algorithm or not, we conduct experiment on the whole testing data, on the partial random testing data, and on the estimated unknown data, respectively. This method is aimed at simulating possible data set used in manufacturing. Therefore, the answer tends to be complete. 

The first threat to construct validity is the suitability of our evaluation metrics. To reduce this risk, we follow the same measurements (ACC, precision, F1, MCC, AUC, stability) as used in other works on failure prediction in production line \cite{carbery2019new,carbery2018bayesian,zhang2016predict,khoza2019comparing,kotenko2019improving,mangal2016using,hebert2016predicting,maurya2016bayesian,huang2019enhancing,moldovan2019time,liu2020adversarial} to measure the quality of the learning model, which reduces the risk of construction validity.

Another threat to construct validity is the threshold $\delta$ ( $\delta = 0.1$ or  $\delta = 0.2$) used in the experiment. In the industry, the threshold value is set according to the production needs. We have performed a hypothesis test for the difference between the two approaches' performances to explain that the difference between FL and CL is within the threshold. The original hypothesis H0 is difference $ > \delta$, and the alternative hypothesis H1 is difference $ < \delta$, where the difference = CL – FL on the same random partial testing data group. Suppose $\alpha$ = 0.05. The p-value of the evaluation metrics for FedSVM vs. SVM and FedForest vs. RForest is shown in Table \ref{tab:1} and Table \ref{tab:2}, respectively. As $alpha > $ p-value for all metrics, the alternative hypothesis H1 is accepted, which means that the difference between CL and FL is less than the threshold $\delta$ ( $\delta = 0.1$ or  $\delta = 0.2$).

\begin{table}[!htbp]
	\centering
	\setlength{\extrarowheight}{2pt}
	\caption{P-value of FedSVM vs. SVM ($\alpha$ = 0.05)}    
	\begin{tabular}{  p{1.4cm}  c  c  c }
		\toprule
		Metrics	& Threshold	& Statistics &	p-value \\
		\midrule
		ACC & 0.1 &	-22.48877 &	5.44616e-41 \\
		Precision	& 0.1	& -83.57313	& 6.33511e-94 \\
		F1 &	0.1	& -17.48523	& 2.45813e-32 \\
		MCC	& 0.2 &	-20.44888 &	1.27870e-37 \\
		AUC	& 0.2 &	-26.63278 &	3.08126e-47 \\
		stability &	0.1	& -5.40692 &	0.00021 \\
		\bottomrule
	\end{tabular}
	\label{tab:1}
\end{table}

\begin{table}[!htbp]
	\centering
	\setlength{\extrarowheight}{2pt}
	\caption{P-value of FedForest vs. RForest ($\alpha$ = 0.05)}    	
	\begin{tabular}{  p{1.4cm}  c  c  c }
		\toprule
		Metrics	& Threshold	& Statistics &	p-value \\
		\midrule
		ACC	& 0.1 &	-34.44843 &	3.16706e-57 \\
		Precision &	0.1 &	-48.40227 &	5.04976e-71\\
		F1 & 0.1 &	-75.16218 &	1.95987e-89 \\
		MCC	& 0.2 &	-12.90324 &	3.11933e-23\\
		AUC & 0.2 &	-16.77851 &	5.26331e-31\\
		stability &	0.1 & -9.83780 & 2.04964e-06\\		
		\bottomrule
	\end{tabular}
	\label{tab:2}
\end{table}

\section{Discussion}
\label{sec:discussion}

Due to data privacy and security, we are facing a bottleneck when applying CL in the industry. In this work, we have adopted two FL algorithms to predict failure in the production line. According to our investigation of existing results of this problem, SVM and RF performed better on Bosch data among CL algorithms that did not consider time series. Therefore, we choose Federated-SVM and Federated-RF in the study. We expect that both will perform as well as the CL algorithms. The results showed that our intuition is not wrong on the given dataset. 

An interesting question is whether FL can achieve the same or similar prediction results as CL. If yes, what are the conditions (e.g., the the dataset's size)? Otherwise, what are the guidelines on when we can use FL in place of CL? As our study is limited to the Bosch dataset from the manufacturing industry, we have conducted a literature investigation to answer this question. According to existing studies \cite{kairouz2019advances,DBLP:conf/iccd/DuanLCTRQL19}, FL's effect is mainly related to the data distribution and data amount. Two important facts are summarized as follows.
\begin{enumerate}
	\item
	When the data is independent and identically distributed, the learning results of FL and CL are similar. For instance, the work \cite{DBLP:conf/hotedge/LuYS19,chen2020federated,DBLP:conf/ispa/SozinovVG18,choudhury2019differential} showed that when the training data is independent and identically distributed (IID), the difference between FL and CL is within 3\%. If the amount of data on each client is small, FL's effect is better than CL because FL expands the number of IID data samples \cite{DBLP:conf/mobicom/IckinVF19,bakopoulou2019federated,roy2019braintorrent,liang2019federated, chen2020fl,suzumura2019towards}.
	\item
	When the training data is unevenly distributed, FL may not achieve the same effect as CL \cite{DBLP:conf/miccai/ShellerREMB18,DBLP:conf/globecom/HuGLM18,DBLP:journals/imwut/FengRSGL20, liu2018fadl, corinzia2019variational, chen2020federated, DBLP:journals/corr/abs-1910-12191, DBLP:conf/ispa/SozinovVG18}. The review article \cite{kairouz2019advances} claimed that if the amount of data on a client is small, the CL model may be better than the FL model trained using data on multi-clients in the case of uneven distribution of training data.
\end{enumerate}

Besides, some works have shown that the effect of FL is also related to encryption algorithms \cite{pfohl2019federated,qi2020fedrec,DBLP:journals/imwut/FengRSGL20,choudhury2019differential,DBLP:journals/tii/LuHDMZ20,DBLP:conf/miccai/LiMXRHZBCOCF19,DBLP:journals/cem/LiSM20}. If the encryption strength increases, the information loss on the data will be worse. So that the effect of FL decreases. The trade-off between data privacy protection and the accuracy of the trained model is still inevitable. It is still considered an open question of how the effect of FL is related to other factors. 

\section{Conclusion}
\label{sec:conclu}
There is increasing attention for federated learning in the manufacturing industry, but few works have studied how federated learning methods perform in practice. In this work, we have conducted an empirical study of comparing FL and CL methods for problem of the failure prediction in the production line. We constructed a horizontal and vertical federated learning scenario based on the Bosch dataset. We implemented FedSVM to compare with SVM in the HFL scenario and designed FedRF to compare it with RF in the VFL scenario. We also designed an experiment process for evaluating the effectiveness of FL and CL algorithms, which can be reused in future studies. Our results reveal that FedSVM and FedRF can replace SVM and RF, respectively, for failure prediction in the Bosch production line. Because the CL algorithm can be replaced by the FL one (1) on the global testing dataset; (2) on the random partial testing dataset; (3) on the estimated unknown Bosch data. The value of each evaluation metric for FL and CL algorithms differ within 0.1. Moreover, the fact that the testing data is heterogeneous enhances the above three conclusions. 

The results of our empirical study reveal that FL can replace CL in some applications and FL is a desirable way of protecting data in the manufacturing industry. More FL techniques can also be investigated in other manufacturing applications. The studies \cite{moldovan2019time,huang2019enhancing,liu2020adversarial} have considered time-series features and showed that the Long Short Term Memory (LSTM) network can improve the prediction result when time-series features are included \cite{huang2019enhancing}. A future work is to design federated LSTM models to compare with centralized LSTM model, and study whether FL can replace CL within this context.

\begin{acknowledgements}
This work was supported by National Key Research and Development Program of China Grant No. 2019YFB1703903, National Natural Science Foundation of China Grant No. 61732019 and No. 61902011, and the Grant No. NJ2018014 of the Key Laboratory of Safety-Critical Software (Nanjing University of Aeronautics and Astronautics), Ministry of Industry and Information Technology.
\end{acknowledgements}

\bibliographystyle{plainnat}       
\bibliography{biblio}

\end{document}